%% file: AA_Main.tex
\documentclass[10pt,twocolumn,letterpaper]{article}

\usepackage{cvpr}
\usepackage{times}
\usepackage{epsfig}
\usepackage{graphicx}
\usepackage{amsmath}
\usepackage{amssymb}
\usepackage{here}
\usepackage{color}
\usepackage{caption}
\usepackage{array}
\usepackage{tabularx}
\usepackage{xcolor}
\usepackage{comment}

\def\shortcite#1{\cite{#1}}

\usepackage[pagebackref=true,breaklinks=true,colorlinks,bookmarks=false]{hyperref}

\def\cvprPaperID{****} 

\newcommand{\claim}[1]{\textcolor{red}{\textbf{[!!!]}}}
\DeclareMathOperator*{\argmin}{argmin}
\begin{document}

\title{Neural Neighbor Style Transfer} 

\author{Nick Kolkin$^{1,2}$ \qquad Michal Ku\u{c}era$^3$ \qquad Sylvain Paris$^2$ \\ Daniel S\'{y}kora$^3$ \qquad Eli Shechtman$^2$ \qquad Greg Shakhnarovich$^1$  \\ \\ Toyota Technological Institute at Chicago$^1$ \qquad Adobe Research$^2$ \\ Czech Technical University in Prague$^3$ }

\twocolumn[{%
\renewcommand\twocolumn[1][]{#1}%
\maketitle
\begin{center}
\centering
\vspace{-0.75cm}
\includegraphics[width=0.9\linewidth]{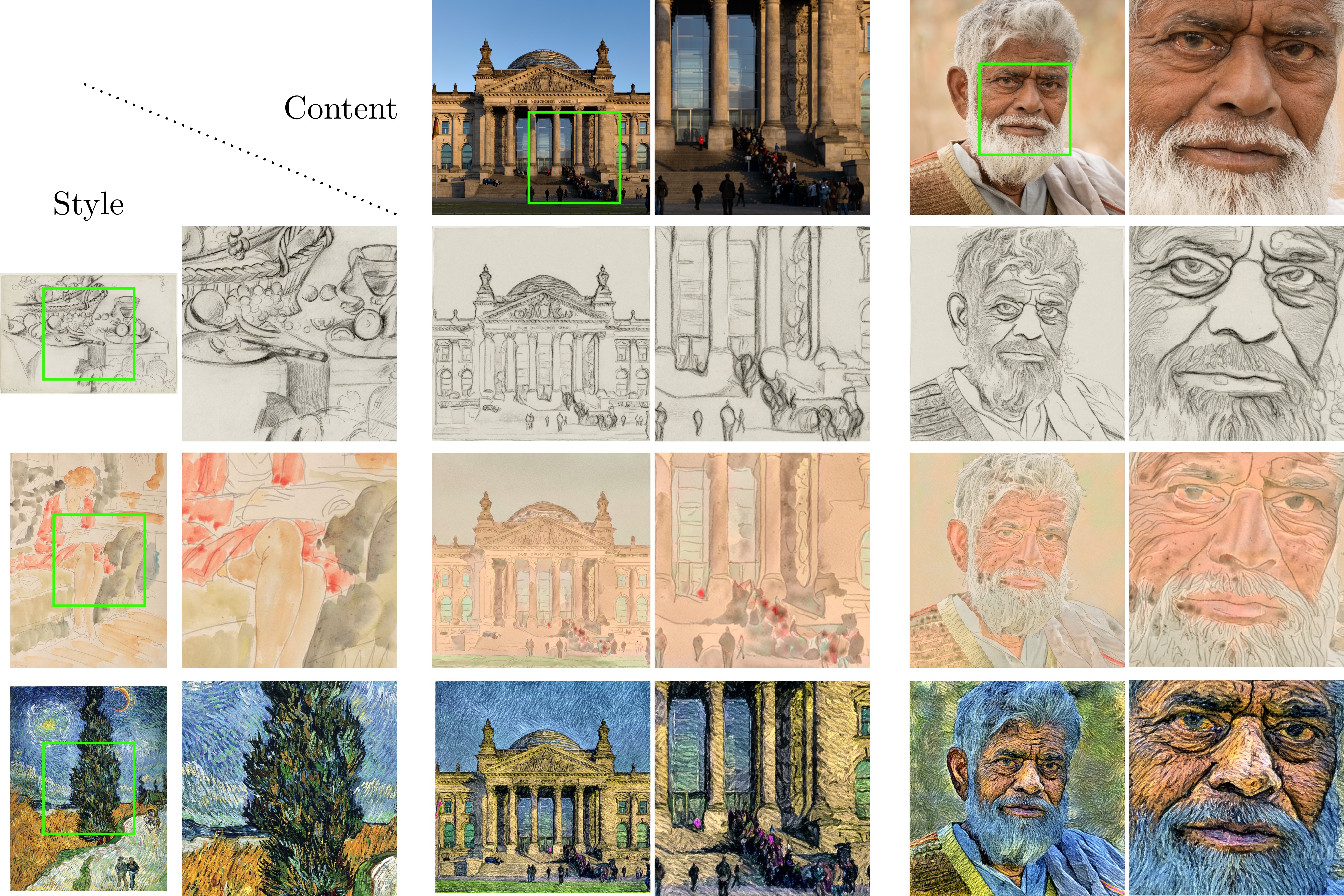}
\captionof{figure}{Examples at 1k resolution of our proposed method, Neural Neighbor Style transfer. Our method synthesizes a stylized output based on rearranging features extracted from the target style using a pre-trained CNN. Synthesis can be implemented either as direct optimization of output pixels (pictured above), or as inference of pixels from features by a learned decoder. Code is available \href{https://github.com/nkolkin13/NeuralNeighborStyleTransfer}{here}}
\label{fig:fig1}
\end{center}
}]

\begin{abstract}
We propose Neural Neighbor Style Transfer (NNST), a pipeline that offers state-of-the-art quality, generalization, and competitive efficiency for artistic style transfer. Our approach is based on explicitly replacing neural features extracted from the content input (to be stylized) with those from a style exemplar, then synthesizing the final output based on these rearranged features. While the spirit of our approach is similar to prior work, 
we show that our design decisions dramatically improve the final visual quality.

There are two variants of our method. NNST-D uses a CNN to directly decode the stylized output from the rearranged style features; it offers similar or better quality than much slower state-of-the-art optimization based methods, and outperforms prior fast feed-forward methods. This version takes a few seconds to stylize a $512\times512$ pixel output, 
fast enough for many applications. NNST-Opt, our optimization-based variant offers higher quality, albeit at lower speed, taking over thirty seconds on the same input size. We compare the stylization quality of both NNST variants with prior work qualitatively and via a large user study with 400 participants which confirms superior performance of our approach. We also demonstrate that NNST can be used for video stylization or extended to support additional guidance and higher output resolution.
\end{abstract}

\input{AB0_intro}

\input{AB1_related}

\input{AB2_method}

\input{AB3_eval}

\input{AB4_conclusion}


\clearpage
%
%
\bibliographystyle{splncs04}
\bibliography{BIB_oat}

\input{AB5_Design}

\end{document}

%% file: AB0_intro.tex

\begin{figure*}
    \centering
    \includegraphics[width=\linewidth]{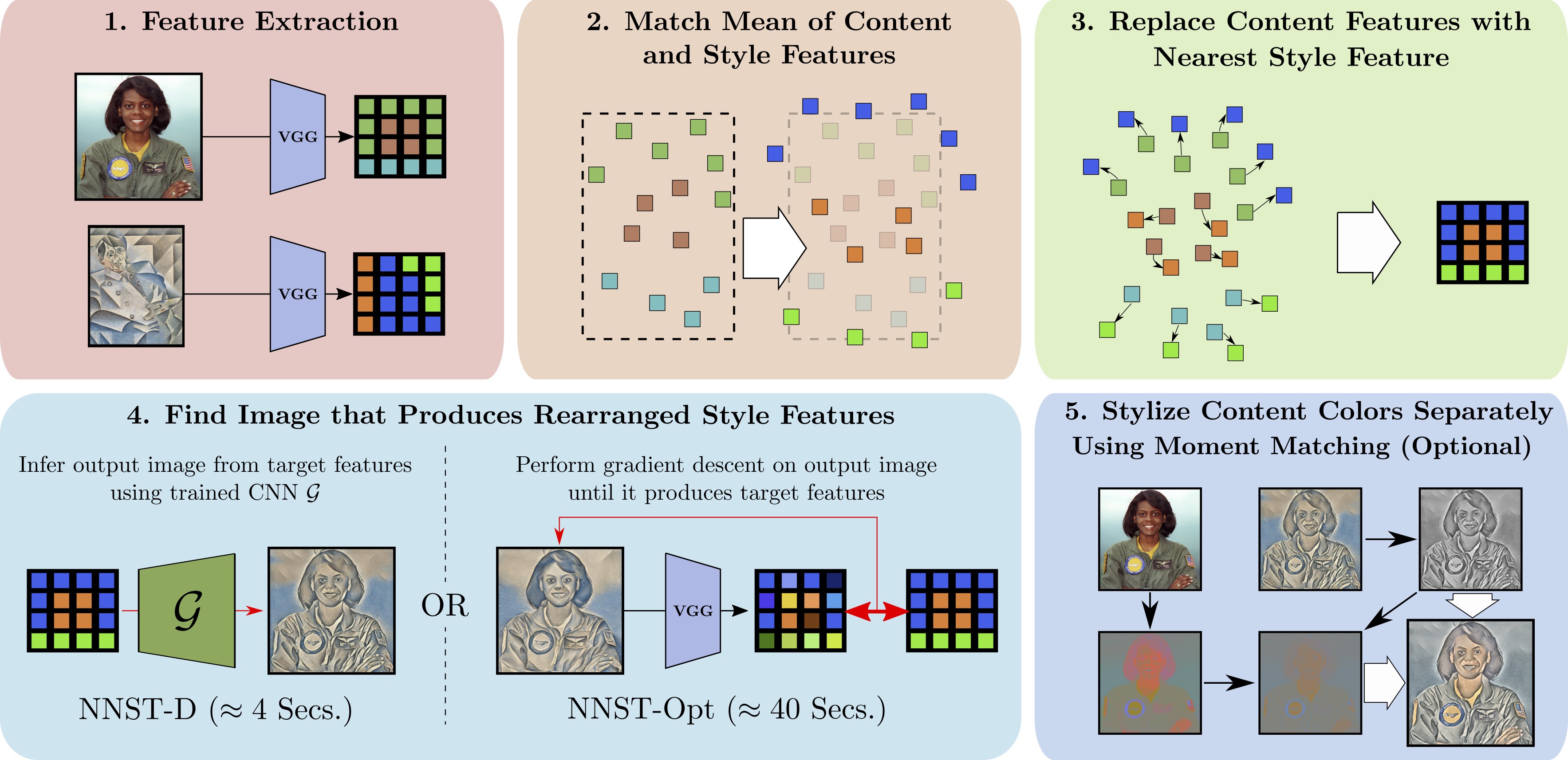}
    \caption{Overview of our method. The fast and slow variants of out method, NNST-D and NNST-Opt, only differ in step 4; mapping from the target features to image pixels. This simplified diagram omits several details for clarity, namely: we apply steps 1-4 at multiple scales, coarse to fine; we repeat steps 1-4 several times at the finest scale; and we only apply step 5 once (optionally) at the very end.}
    \label{fig:overview}
\end{figure*}

\section{Introduction}

Thanks to recent advances in neural style transfer pioneered by Gatys et al.~\cite{gatys2016image} input photos can be turned into a stylized artwork that has a similar content while mimicking distinctive visual features of the style exemplar. The great advantage of neural methods is that they usually do not require additional information to perform the stylization, only the content and the style image need to be specified by the user. In contrast, traditional patch-based approaches~\cite{hertzmann2001image, Fiser16} require additional guidance prepared manually or automatically generated to achieve convincing results. Due to this requirement, patch-based methods usually expect a specific target domain (e.g., example-based stylization of 3D models~\cite{Fiser16} or faces~\cite{Fiser17}) where automatic computation of guidance channels is feasible. Or in a more generic scenarios such as video stylization, they need style exemplars to be precisely aligned with the target content to establish a one-to-one mapping between the content and style~\cite{Jamriska19}. However, a key benefit of patch-based techniques is their ability to accurately preserve visual aspects of the given artistic media, including the notion of individual brush strokes or other details vital to matching the style of the original artwork. This is possible due to the non-parametric nature of patch-based methods, i.e., their ability to generate stylized images by stitching a set of smaller bitmap chunks copied directly from the style exemplar. In contrast, state-of-the-art neural techniques such as~\cite{kolkin2019style} use parametric synthesis where the output image is generated by directly optimizing the values of the output pixels. In this scenario it is difficult to compete with the ability to reuse exact patches from the style exemplar. Historically this has resulted in a visual gap between the style preservation quality of neural and patch-based techniques. On the other hand, besides patch-based method's need for guidance, relying on patches also limits the output's degrees of freedom since such a method can only copy and shift larger chunks of the original style exemplar. To alleviate such drawback, recent style transfer methods propose to combine aspects of both directions~\cite{li2016combining, liao2017visual, gu2018arbitrary, Texler20-SIG}, i.e., use either non-parametric approach in the space of neural features and then perform parametric synthesis to obtain the final image or vice versa. For instance, Deep Image Analogy of Liao et al.~\cite{liao2017visual} delivers results approaching quality of patch-based synthesis~\cite{Fiser17} without the need to prepare extra guidance, however, it still requires the style image to have a similar content as the target image.

In this work our aim is to further elevate the visual quality of neural style transfer without requiring extra guidance or domain specific style exemplars. We adopt the combined approach to style transfer, i.e., we construct target neural features by rearranging features of the style image and then we find an image which produces these features either by optimization or a learned decoder network. Our main contribution is hidden in the feature construction phase. Here we demonstrate how to find a good balance between the overall visual diversity, stylistic fidelity, and content preservation by combining three improvements: (1)~zero-centering the content and style features, (2)~using the cosine distance for measuring feature similarity, (3)~performing feature splitting during the nearest-neighbour matching phase which is computed in every iteration of the target feature construction, and most importantly (4) matching individual neural feature vectors rather than patches as proposed in \cite{chen2016fast, li2016combining}. In our evaluations and ablation study we demonstrate how those four components enable significant improvement over current state-of-the-art. Qualitatively our results more accurately capture the texture of the target media than prior work, particularly when seen at high resolution.

%% file: AB1_related.tex
\section{Related Work}\label{sec:related}
Example-based style transfer, non-photorealistic rendering guided by a
single piece of artwork, is a widely studied image synthesis
task. Approaches like our own, in which image synthesis is guided by
explicit matches between spatially localized content and style
features, can be traced to \cite{hertzmann2001image} and related work
on texture synthesis
\cite{efros2001image,efros1999texture,wei2000fast}. Recent patch-based style transfer algorithms have focused on producing high quality outputs when additional guidance is available, either in the form of boundary annotations~\cite{Lukac13,Lukac15}, or a set of custom tailored guiding channels~\cite{benard2013stylizing,Fiser16,Fiser17,Jamriska19}. These methods, however, are applicable only in scenarios when such guidance can be provided either manually or computed automatically. Such requirements are usually violated in our arbitrary style transfer setting where there are no assumption imposed on the input style image.

\textbf{Optimization-based Neural Style Transfer:} Recent years have marked a significant departure from this line of work, building off techniques pioneered by Gatys et al.~\cite{gatys2016image} in their `A Neural Algorithm of
Artistic Style'. This work contributed in two key ideas. Firstly, it shows how to leverage features extracted by a convolutional neural network pre-trained for
image classification tasks (typically VGG~\cite{simonyan2014very}) as a way how to measure high-level compatibility between two images. The
second idea was direct optimization of the output's pixels using gradient
descent to simultaneously minimize a `style loss' (matching statistics derived from the features of the VGG artwork), and
a `content loss' (minimizing deviation from the content image's features). Many follow-up works~\cite{gatys2016image,berger2016incorporating,li2016combining,risser2017stable,mechrez2018contextual,gu2018arbitrary,kolkin2019style} proposed alternative style losses or introduced content loss invariant to translations and rotoreflections in the feature space~\cite{kolkin2019style}.

In the arbitrary style transfer setting, these approaches performs better than patch-based techniques, nevertheless, methods that use a content and style loss, both based on VGG features, face a fundamental challenge. In general it is impossible to match the statistics of the style features (satisfying the style loss) while keeping the content features unchanged (satisfying the content loss). Even if the content loss is invariant to rotoreflections~\cite{kolkin2019style}, matching the second order statistics captured by the simplest original style loss~\cite{gatys2016image,li2017demystifying} generally requires an affine transformation (of which rotoreflections are a subset, and therefore do not provide sufficient invariance). In practice we observe that content losses based on deep VGG layers tend to cause photographic high frequencies of the original content to bleed into the final output.

To avoid such a drawback, our proposed algorithm borrows the idea of guided patch-based synthesis to explicitly match content and style features, however, instead of working directly on image patches it uses features extracted from the responses of pre-trained VGG network.

\begin{figure*}[tp]
\centering
\includegraphics[width=\linewidth]{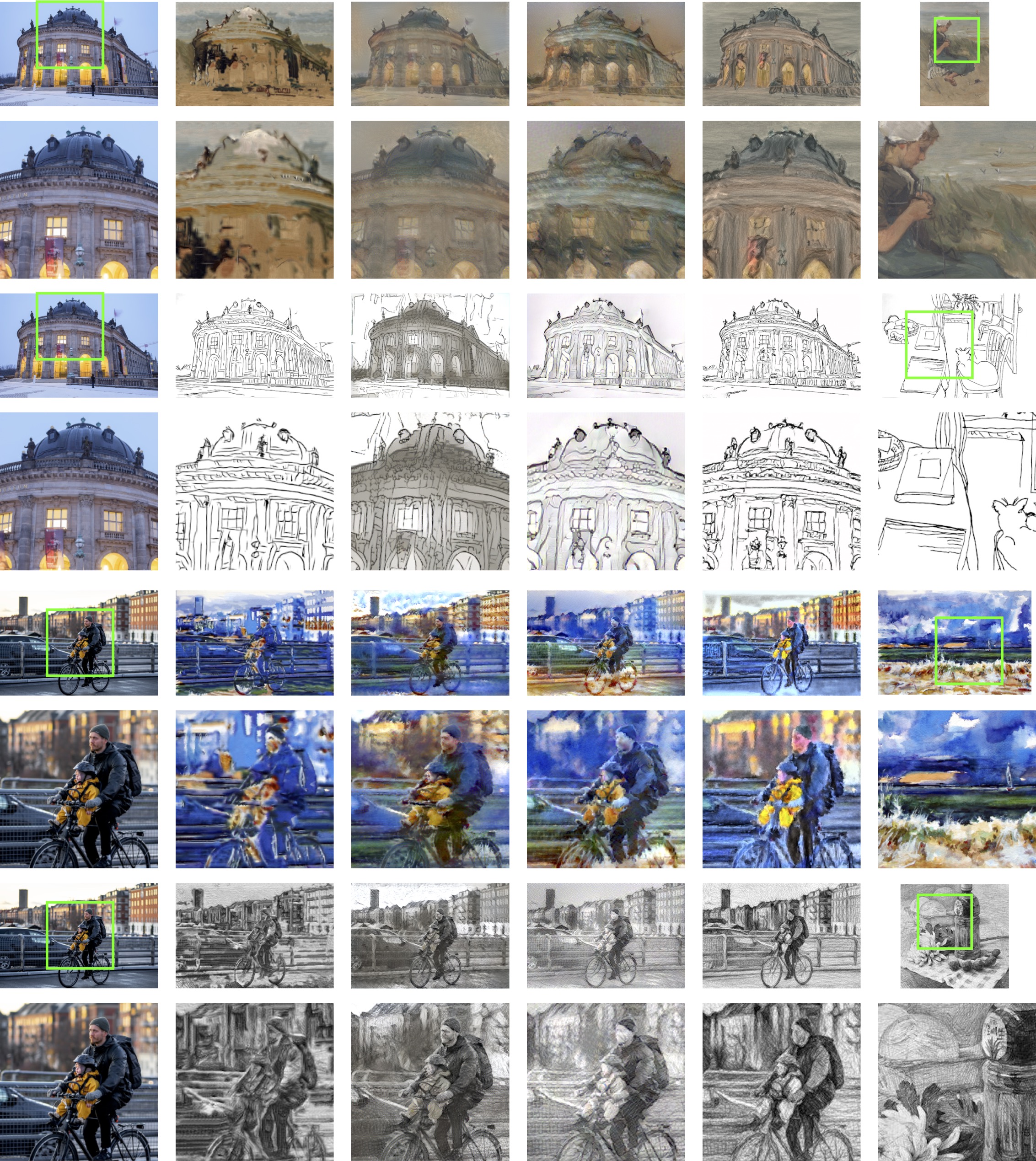}
\begin{tabularx}{\linewidth}{>{\centering\arraybackslash}X>{\centering\arraybackslash}X>{\centering\arraybackslash}X>{\centering\arraybackslash}X>{\centering\arraybackslash}X>{\centering\arraybackslash}X}
Content & Deep Image Analogies \cite{liao2017visual} & Gatys \cite{gatys2016image} & STROTSS \cite{kolkin2019style} & NNST-Opt (Ours) & Style 
\end{tabularx}
\caption{Qualitative comparison between NNST-Opt and the top three optimization based methods from our user study, using oil painting, ink, watercolor, and pencil styles. Below each input and result is a zoomed-in portion of the image. While all neural methods to date fail to entirely capture many styles' long range correlation of textural features and high frequency details, our results are dramatically closer than prior work.}
\label{fig:qual_slow}
\end{figure*}

Similar approach was used previously in CNNMRF~\cite{li2016combining} that replaces style loss of~\cite{gatys2016image} with minimizing each patch of content feature's distance from its nearest neighbor patch of style features under the cosine distance. However, it also regularizes its outputs using the content loss proposed in~\cite{gatys2016image}, leading to the fundamental tension outlined above. Other approaches tried to explicitly construct a set of target features without considering content loss. In~\cite{chen2016fast,li2016combining} overlapping feature patches from the style image are averaged. This approach preserves content well, however, at the cost of fidelity to the target style. Gu et al.~\cite{gu2018arbitrary} add a soft constraint on the matches found by nearest neighbors that each style vector may only be used at most~$k$ times. Nevertheless, this constraint can be overly restrictive and lead to content distortion artifacts. Liao et al.~\cite{liao2017visual} proposed a coarse-to-fine strategy for finding feature correspondences between the content and style image. Their approach produces excellent results on style-content pairs with matching semantics and similar poses (and based on our user study, ink based styles using sparse lines), but in general does not work for more disparate style-content pairs (see Figure~\ref{fig:qual_slow}).
Finally, Texler et al.~\cite{Texler20-CAG} propose a method based on explicit neural feature matching to guide patch-based synthesis. However, since the final output can only be a mosaic of small bitmap chunks taken from the original style image the flexibility to express different target content can be fairly limited.


\begin{figure*}[tp]
\centering
\includegraphics[width=\linewidth]{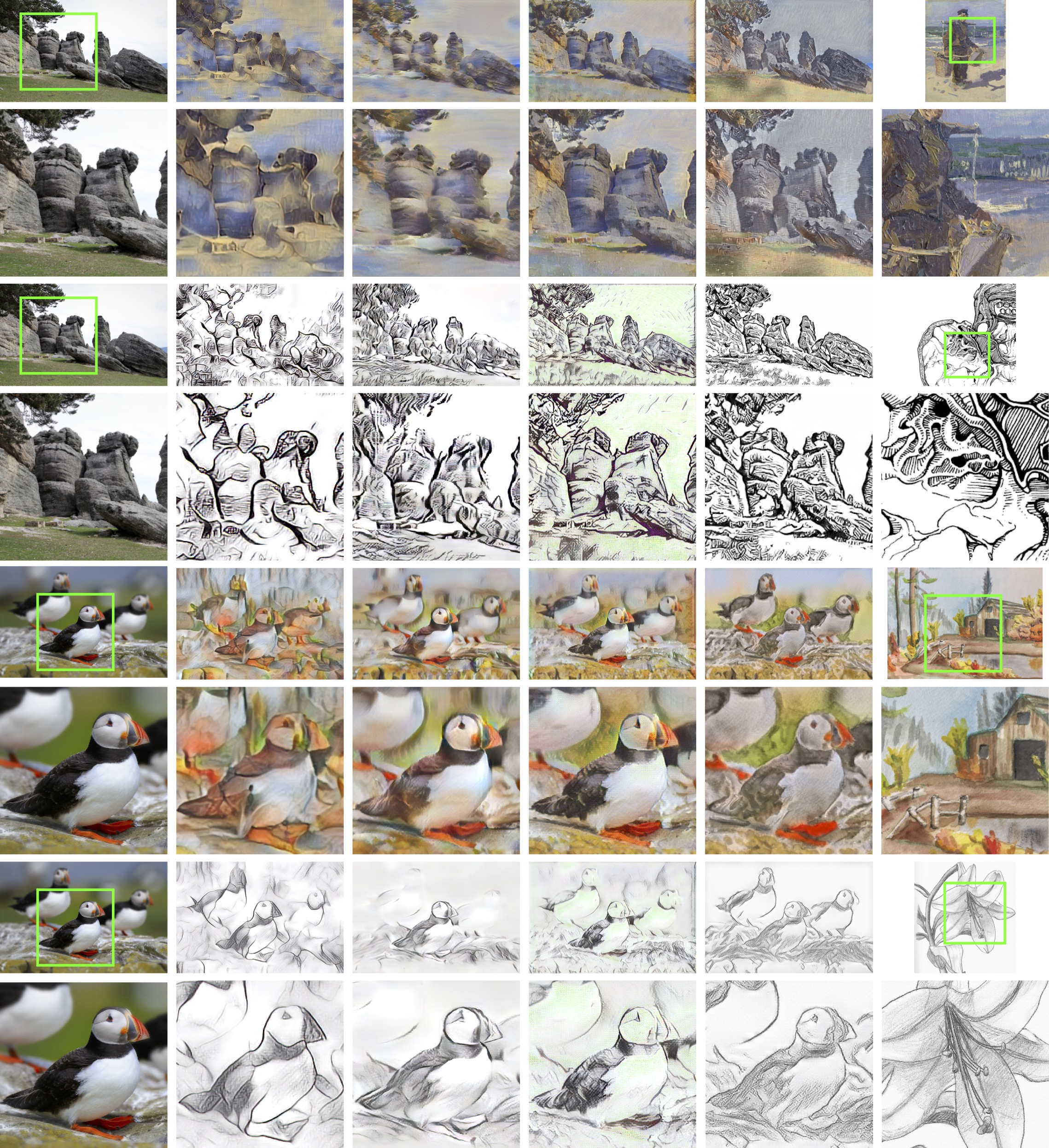}
\begin{tabularx}{\linewidth}{>{\centering\arraybackslash}X>{\centering\arraybackslash}X>{\centering\arraybackslash}X>{\centering\arraybackslash}X>{\centering\arraybackslash}X>{\centering\arraybackslash}X}
Content & WCT \cite{Li17} &  MST \cite{zhang2019multimodal} & ArtFlow \cite{an2021artflow} & NNST-D (Ours) & Style 
\end{tabularx}
\caption{Qualitative comparison between NNST-D and the top three feed-forward methods from our user study, using oil painting, ink, watercolor, and pencil styles. Below each input and result is a zoomed-in portion of the image. While all neural methods to date fail to entirely capture many styles' long range correlation of textural features and high frequency details, our results are dramatically closer than prior work.}
\label{fig:qual_fast}
\end{figure*}

\textbf{Feed-Forward Neural Style Transfer: } 
A shortcoming of optimization-based style transfer methods is their computational overhead. To address this a large body research has focused on training feed-forward networks which can perform style transfer quickly~\cite{Huang17,Li17,chiu2019understanding,yao2019attention,park2019arbitrary,zhang2019multimodal, Chiu2020Iterative,an2021artflow,sheng2018avatar}. Many of these techniques use closed-form modifications of features extracted from the content image based on the first and second order statistics of the style features~\cite{Huang17,Li17,chiu2019understanding}.  Sheng et al.~\cite{sheng2018avatar} propose a 'style decorator' to hallucinate alternate content features more amenable to stylization. Chiu et al.~\cite{Chiu2020Iterative} replace single stage moment matching with an efficient method for iterative stylizing the content features via analytical gradient descent. Yao et al.~\cite{yao2019attention} uses self-attention to improve content preservation in perceptually important regions and Park et al.~\cite{park2019arbitrary} use self-attention to distribute localized style features according to the content's self-similarity. Zhang et al.~\cite{zhang2019multimodal} improve stylization quality by relaxing the assumption that the feature distributions are uni-modal, and matching clusters of features using a graph cut. Recently, An et al.~\cite{an2021artflow} uses normalizing flows to stabilize style transfer, improving stylization quality and preventing results from changing after multiple rounds of stylization. However, because these methods either learn to extract information from the style image, or only have access to limited feature statistics, they often struggle to reproduce distinctive elements of the target style. This gives them a disadvantage relative to optimization-based methods, which 'learn from scratch' for each new input image pair, and consequently the visual quality of the feed-forward results is lower. By taking as input a large tensor of rearranged style features extracted by pre-trained VGG, NNST-D has direct access to much richer information about the target style.

\textbf{Targeted Feed-Forward Neural Style Transfer:} Another common scenario, which we do not address in this work, is when the styles of interest are known beforehand, and a neural network can be pre-trained to produce stylizations of the predetermined type(s)~\cite{johnson2016perceptual,Zhu17a,junginger2018unpaired,sanakoyeu2018style,kotovenko2019content, svoboda2020two}. These methods are very fast, and can produce high quality results. However, they require enough training data for a given style, and must be retrained for new styles.



%% file: AB2_method.tex

\section{Neural Neighbor Style Transfer}\label{sec:ot}
We implement our method using the PyTorch framework \cite{NEURIPS2019_9015}. The feed-forward variant of our method, NNST-D takes 4.5 seconds to process a pair of 512x512 content/style images. Our optimization based variant, NNST-Opt, takes 38 seconds to process the same input. Timing results are based on an NVIDIA 2080-TI GPU.
\subsection{Feature Extraction}
Our pipeline relies on a pre-trained feature extractor $\Phi(x)$, where $x$ is an RGB image.  $\Phi(x)$ extracts the hypercolumns \cite{mostajabi2015feedforward,hariharan2015hypercolumns} formed from the activations produced for convolutional layers in the first four blocks of pre-trained VGG16 \cite{simonyan2014very} when $x$ is passed in. We use bilinear interpolation on activations from all layers to give them spatial resolution equal to one quarter of the original image. For an image with height H, and width W, this yields an image representation  $\Phi(x) \in \mathbb{R}^{\frac{H}{4} \times \frac{W}{4} \times 2688}$. Generally we consider style to be rotation invariant, and to reflect this we extract features from the style image rotated at $0^{\circ}, 90^{\circ},180^{\circ}$ and $270^{\circ}$ in all experiments.

\subsection{Feature Matching} \label{sec:HM}
\begin{figure}
    \centering
    \includegraphics[width=\linewidth]{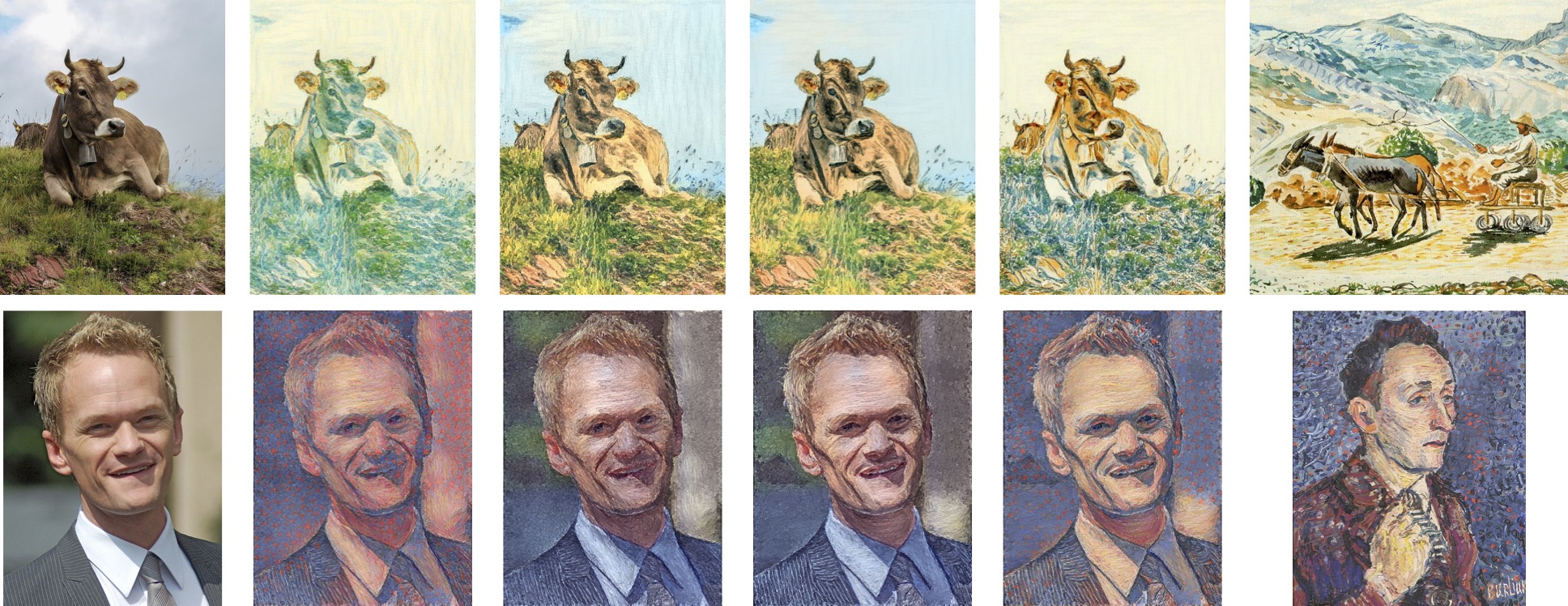}
        \begin{tabularx}{\linewidth}{>{\centering\arraybackslash}X>{\centering\arraybackslash}X>{\centering\arraybackslash}X>{\centering\arraybackslash}X>{\centering\arraybackslash}X>{\centering\arraybackslash}X}
    \hspace{-0.2cm}Content & \hspace{-0.3cm}(a) &  \hspace{-0.5cm}(b) & \hspace{-0.6cm} (Ours) & \hspace{-0.7cm}(d) & \hspace{-0.3cm}Style
    \end{tabularx}
    \caption{Demonstration of the affect of zero-centering features before nearest-neighbor matching. In (a) there is no zero-centering and no color processing, resulting in lower quality feature pairings that lead to more homogeneous colors and worse content preservation. (b) largely fixes (a) by adding color processing, although the features of the face are poorly defined. (Ours) is the default setting of NNST which uses zero-centering and color-correction, producing good results in both cases. (d) is the same as (Ours) but with no color processing, this setting is less reliable and more prone to introducing visual errors than (Ours), but when colors are mapped correctly the results can be stunning and zero-centering makes this more frequent. As zero-centering is efficient and we never observe it to hurt results, we always include it.}
    \label{fig:mu}
\end{figure}

The core steps of our pipeline are outlined in Figure \ref{fig:overview}. We extract features from the style image and content image (1) zero-center the content features and style features (2). Then use nearest-neighbors matching under cosine distance (3) to replace each content feature (hypercolumn) with the closest style feature. If the content image $C$ is of size $H_c \times W_c$, and style image $S$ is of size $H_s \times W_s$, this yields a new target representation for our stylized output $T \in \mathbb{R}^{\frac{H_c}{4} \times \frac{W_c}{4} \times 2688}$ where the feature vector $T_{i}\in\mathbb{R}^{2688}$ at each spatial location is derived from the original style image, or a rotated copy. For simplicity let $\Phi'(x)$ be the function extracting features from $x$ and its rotations, where an individual feature vector (from any spatial location in any rotation) can be indexed as $\Phi'(x)_j$. Formally:

\begin{align}\label{eq:match}
    T_i &= \argmin_{\Phi'(S)_j} D\Big(\Phi(C)_i - \mu_C, \hspace{0.2cm}\Phi'(S)_j - \mu'_S \Big)
\end{align}

Where $D$ is the cosine distance, $\mu_C$ is the average feature extracted from the content image, and $\mu'_S$ is the average feature extracted from the style image and its rotated copies. While mean subtraction does not have a huge impact when using our color post-processing, it extremely important without it, enabling some stunning results in cases where the content and style are well matched (Figure \ref{fig:mu}).

\begin{figure}
    \centering
    \includegraphics[width=\linewidth]{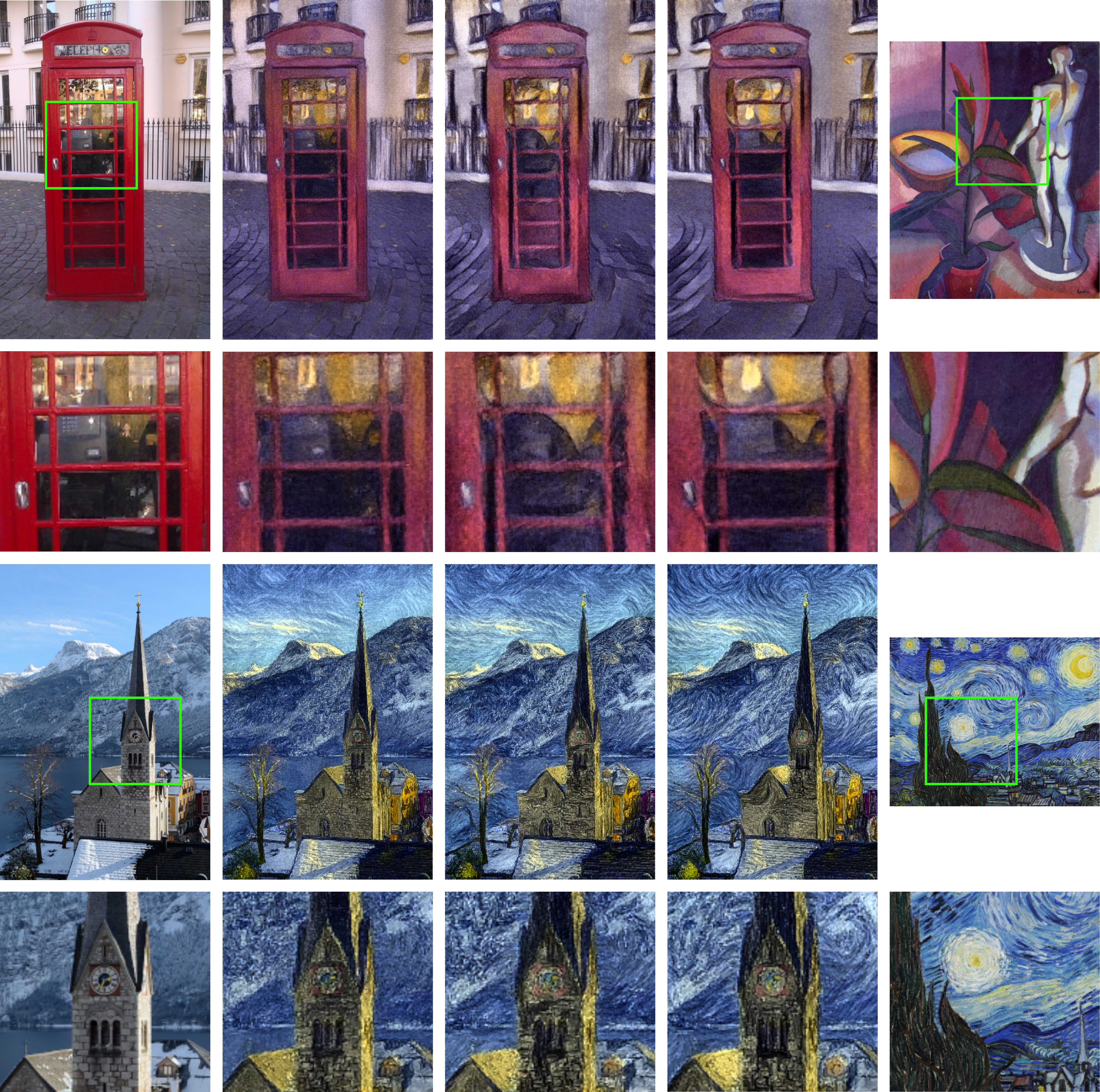}
    \begin{tabularx}{\linewidth}{>{\centering\arraybackslash}X>{\centering\arraybackslash}X>{\centering\arraybackslash}X>{\centering\arraybackslash}X>{\centering\arraybackslash}X}
    Content & (a) &  (b) & (c) & Style
    \end{tabularx}
    \caption{Demonstration of the effect of our final feature splitting phase (c). (a) is our result without any final phase, content is well preserved, but too many photographic details are preserved and brushtrokes in the 2nd row are poorly defined. (b) mimics our feature splitting phase, recomputing feature matches many times, but uses hyper-columns instead of computing matches separately for each layer. This leads to unnecessary loss of content details and muddier high frequencies relative to (c).}
    \label{fig:regime}
\end{figure}

In sections \ref{sec:d} and \ref{sec:opt} we describe our feed-forward and optimization based methods for recovering image pixels from $T$ (choosing between these differentiates between NNST-D and NNST-Opt). In the main loop of our pipeline  we produce stylizations at each scale, coarse to fine, and each result is used to initialize the next scale. Throughout this process we match hypercolumns wholesale, and keep the $T$ unchanged throughout the synthesis process at a particular scale. While this is efficient (since $T$ need only be computed once per scale, and computing a single large distance matrix is well suited to GPU parallelism), and the result roughly captures many aspects of the target style, stopping at this point leads to images that are fail to capture the high-frequencies of the target style (Figure \ref{fig:regime}).

We believe that this effect is due to incompatible hypercolumns, which are not adjacent in the original style, being placed next to each other in $T$. Because these features have overlapping receptive fields, the output is optimized to produce the average of several features (each taken from a different region of the style) at a single output location. This manifests visually as a 'washed out' quality, an issue noted by prior work in style transfer \cite{gu2018arbitrary}, and other patch-based synthesis work~\cite{kaspar2015self,Jamriska15,Fiser16}.

We find that these issues can be largely resolved by a final phase where the the feature matching process is less constrained, a similar solution to one used by Luan et al.~in the image compositing~\cite{luan2017deep}. In this final phase, which we call 'Feature Splitting', matches are computed for each layer separately, resulting in $T$ consisting of novel hypercolumns where features at different layers are mixed and matched from different locations/rotations of the style image. Unlike~\cite{luan2017deep} we do not compute matches only once, we recompute them after every update to the output image. In this phase features are matched relative to the current output, rather than the initial content. When using our learned decoder $\mathcal{G}$ to synthesize, this amounts to feeding the output back into the same network as 'content' five times (recomputing $T$ each time). When directly optimizing the output image, this amounts to recomputing $T$ using the current output as the 'content' after each Adam update. 

\subsection{Neural Network Decoder (NNST-D)} \label{sec:d}
\subsubsection{Architecture} The decoder $\mathcal{G}$ takes as input the target representation $T \in \mathbb{R}^{\frac{H_c}{4} \times \frac{W_c}{4} \times 2688}$. $T$ is then fed into 4 independent branches, each responsible for producing one level of a 4-level laplacian pyramid parameterizing the output image. Each branch has virtually the same architecture (but separate parameters), consisting of five 3x3 convolutional layers with leaky relu \cite{maas2013rectifier} activations (except the last layer, which is linear), and a linear residual 3x3 convolution \cite{he2016deep} directly from $T$ to the branch's output. All intermediate hidden states have 256 channels. The four branches differ only in number of output channels, having $4^2\times3$, $2^2\times3$, $3$, and $3$ output channels respectively. Transposed convolutions are applied to the first two branches to trade off channel depth for resolution (resulting in one $H_c \times W_c \times 3$ output and one $\frac{H_c}{2} \times \frac{W_c}{2} \times 3$ output). The third branch is not altered (resulting in a $\frac{H_c}{4} \times \frac{W_c}{4} \times 3$ output), and the output of the fourth branch is bilinearly downsampled by a factor of two (resulting in a $\frac{H_c}{8} \times \frac{W_c}{8} \times 3$ output). The final output image is synthesized by treating the output of the four branches as levels of a laplacian pyramid and combining them appropriately. 

\subsubsection{Training} We train our model using MS-COCO as a source of content images, and Wikiart as a source of style images, matching the training regime of \cite{zhang2019multimodal, an2021artflow, park2019arbitrary}. Content/Style training pairs are randomly sampled independently from each dataset. For each input pair, two outputs are generated during training, a reconstruction of the style image and a style transfer:
\begin{align}
    \hat{S}  &= \mathcal{G}(\Phi(S)) \\
    \hat{C_S} &=\mathcal{G}(T)
\end{align}
Recall that an intermediate output of $\mathcal{G}$ is a laplacian pyramid that is collapsed to form the final output image. Let the levels of this pyramid for the style reconstruction be $\hat{S}_{1..4}$, let a 4-level laplacian pyramid constructed directly $S$ be $S_{1..4}$, let $P_i$ be the number of pixels at level $i$. These are used to compute the reconstruction loss:
\begin{equation}
    \mathcal{L}_r = \sum_{i=1}^4 \frac{\|S_i - \hat{S}_i\|_1}{P_i}
\end{equation}
We do not know what the pixels of the style transferred result should be, so we instead optimize this output using a cycle loss on $T$:
\begin{equation}
    \mathcal{L}_{cycle}=\frac{16}{H_CW_C}\sum_{i=0}^{\frac{H_CW_C}{16}}D \Big(T_i, \hspace{0.2cm}\Phi(\hat{C_S})_i \Big)
\end{equation}
Where $D$ computes the cosine distance, and $i$ indexes over the spatial indexes of $T$ and $\Phi(\hat{C_S})$ (which are both a quarter of the original resolution of $C$). To further improve the 'realism' of our results and encourage better capturing the target style we also employ the adversarial patch co-occurrence loss proposed by Park et al. \cite{park2020swapping}:
\begin{equation}
    \mathcal{L}_{adv}= -\log \mathcal{D}(\Theta^{(4)}(S), \Theta^{(1)}(\hat{C_S}))
\end{equation}
Where $\Theta^{(k)}(x)$ is a function that extracts $k$ random patches of size $\frac{\max(H,W)}{8}$ from $x$, and $H,W$ are the height and width of $x$ respectively. $\mathcal{D}$ is a discriminator that evaluates a single patch, conditioned on 4 patches extracted from the style image. We use the same discriminator architecture and discriminator training described in \cite{park2020swapping}, where further details can be found. We fit the parameters of our model, $\theta_\mathcal{G}$ to minimize the full objective:

\begin{align}
    \min_{\theta_\mathcal{G}} \mathbb{E}_{C\sim\mathbb{P}_C,\,S\sim\mathbb{P}_S}\Big[\mathcal{L}_r+ \mathcal{L}_{cycle} + \mathcal{L}_{adv} \Big]
\end{align}

\begin{figure}[htp]
    \centering
    \includegraphics[width=\linewidth]{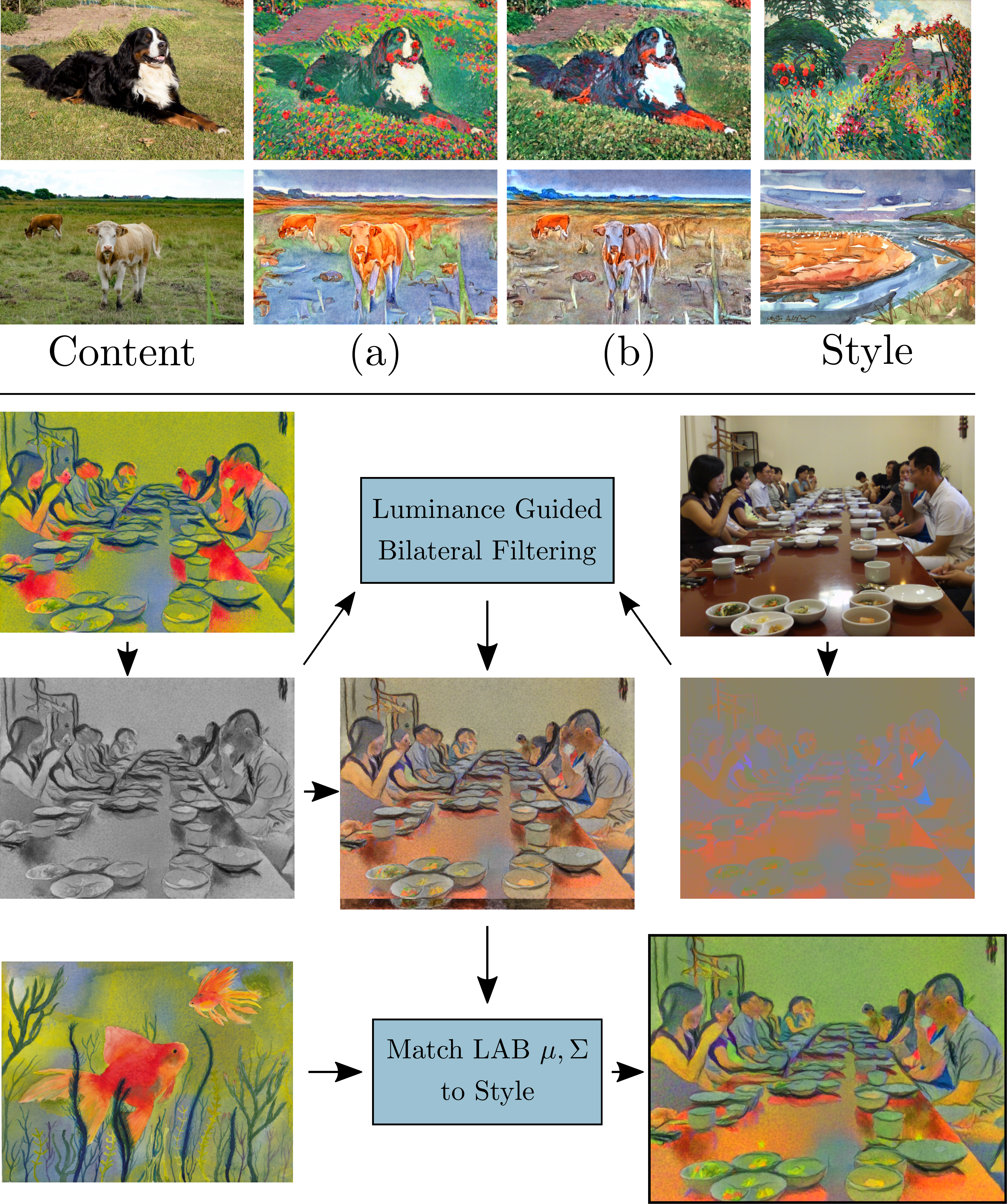}
    \caption{In the first two rows we give examples of NNST-Opt with color processing (b) and without (a). Color processing helps fix common content preservation errors due to features with inconsistent colors being matched to the same object or region. Below these examples we outline our the color processing procedure. First the stylized luminance produced by optimization or the decoder is extracted. Then it is used as a guide for bilateral filtering on the original content's AB channels, this aligns the boundaries of colored regions to the stylized L channel. After combining this with the stylized L channel with the filtered AB channels we use simple moment matching to align the output's color distribution with the style.}
    \label{fig:color}
\end{figure}

Where $\mathbb{P}_C$, $\mathbb{P}_S$ are the distributions of content and style training images respectively. A separate decoder is trained for each output scale (64, 128, 256, 512 pixels on the long side). Training converges fairly quickly, and we use models trained for a three epochs on MS-COCO (Wikiart images are sampled independently with replacement for each MS-COCO example). We train using a batch size of 4, and the Adam optimizer \cite{kingma2019method} with parameters $\eta=2e^{-3}, \beta_1=0.0, \beta_2=0.99$. The same set of decoder models (four total, one for each scale) are used in all experiments.

\subsection{Image Optimization (NNST-Opt)} \label{sec:opt}
While performing style transfer using $\mathcal{G}$ is fast, there are many cases where optimizing the output image directly produces sharper results with fewer artifacts. Given our target features $T$ and feature extractor $\Phi$, we find output image $x$ by minimizing following objective:
\begin{equation}\label{eq:main}
    \min_x \hspace{0.3cm} -\frac{1}{P} \sum_{i=0}^{P-1}\cos\left(\Phi(x)_i,T_i\right)
\end{equation}
where $P=W_cH_c/16$, the number spatial locations in $\Phi(x)$ and $T$. 

Equation \ref{eq:main} is minimized via 200 updates of $x$ using Adam \cite{kingma2019method} with parameters $\eta=2e^{-3}$, $\beta_1=0.9$, and $\beta_2=0.999$. To allow the average color of large regions to quickly change within 200 updates, we parameterize $x$ as a laplacian pyramid with 8 levels.

\subsection{Color Post-Processing}

We find that a common source of perceptual errors in the outputs produced by NNST-D, NNST-Opt, and other methods, is when a region with a single color in the original content is mapped to multiple colors in the output (see the first two rows of Figure \ref{fig:color}). However, after converting our outputs to Lab colorspace, the luminance channel generally matches the target style well, and is free of artifacts. This motivates our post-processing step, which is to take the luminance generated by NNST-D or NNST-Opt, but discard the AB channels and replace them with the AB channels of the original content. In order to match the content's original AB channels to the generated L channel we perform bilateral filtering \cite{Tomasi1998BilateralFF,paris2009bilateral} on the AB channels guided by the L channel. Then we match the mean and covariance of color distribution formed by the output's L channel and filtered AB channels to the style's color distribution. (see the bottom section of Figure \ref{fig:color}). This is similar to the color control proposed in \cite{gatys2017controlling}, 
but allows stylizations which more closely match the palette of the target style.

In the vast majority of cases this post-processing step improves results, however we have observed a few scenarios where it does not. 
First, for apparently monochrome styles (e.g., pencil or pen drawings), visually imperceptible color variations in the style can distort the second-order statistics used in moment matching, leading to results where desaturated colors are visible. Fortunately in these cases the unprocessed results of NNST are typically monochrome, and we detect this situation by examining the maximum between the variance of the A and B channels, then not applying the color processing if the value is below a threshold (we find 4e-5 to work well).

Second, this post-processing step can prevent the output from matching distinctive color features of the style, such as local color variations as in the dots of pointillism, the abrupt color shifts within objects of cubism, or the limited multi-modal palettes used in some artwork (see Figures \ref{fig:mu} and \ref{fig:degree}). Third, colorful styles with a large white background can lead to moment matching causing over-saturation. In these cases it can be better to not use this post-processing step.

Fortunately, our post-processing step is simple and computationally efficient, taking less than a millisecond in our PyTorch implementation. In a practical setting it is essentially free for users to generate results both with and without the post-processing, and choose the one which best suits their needs.

\subsection{Control of stylization degree}
\begin{figure}
    \centering
    \includegraphics[width=\linewidth]{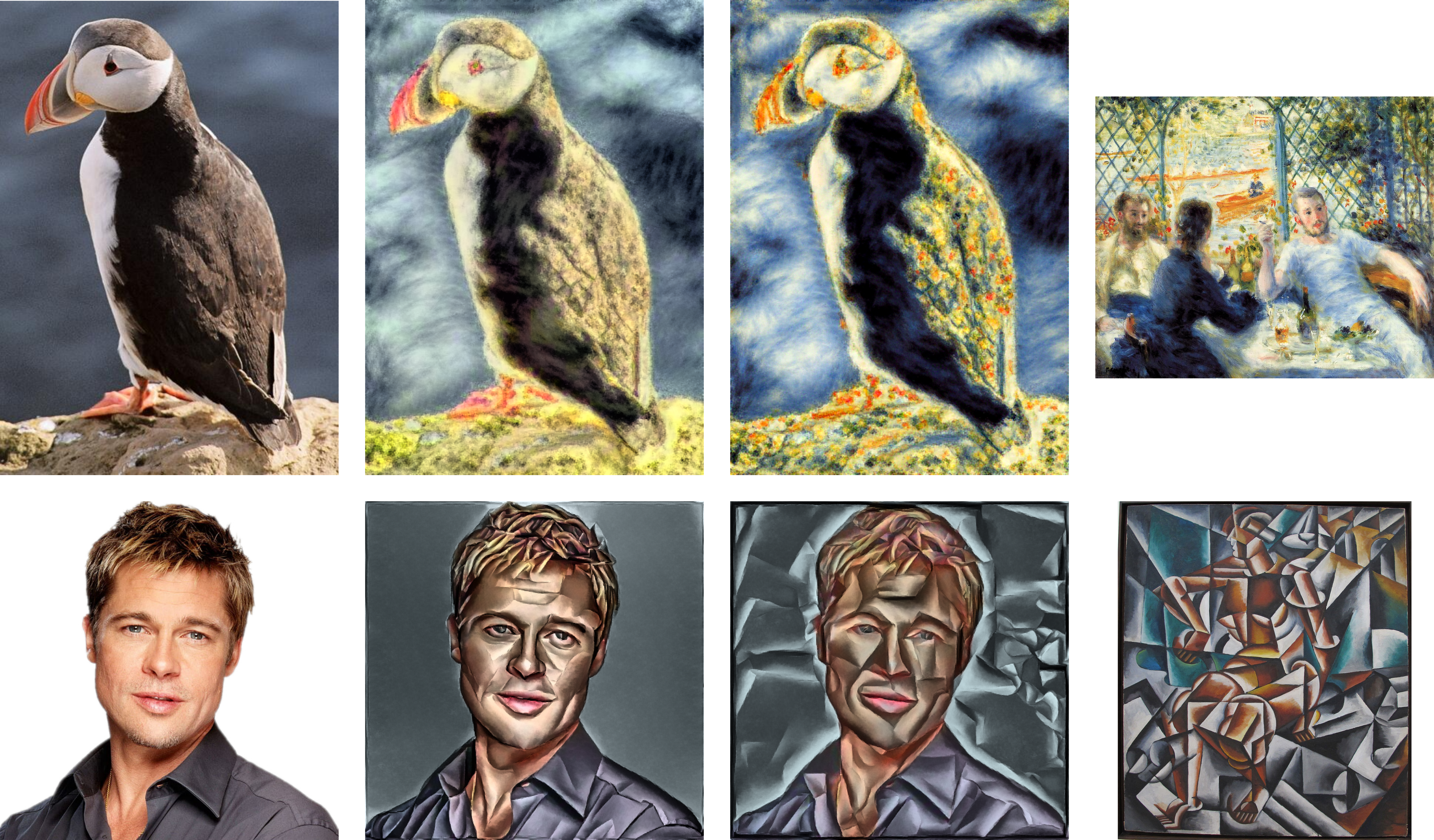}
    \begin{tabularx}{\linewidth}{>{\centering\arraybackslash}X>{\centering\arraybackslash}X>{\centering\arraybackslash}X>{\centering\arraybackslash}X}
    Content & $\downarrow$ Stylization &  $\uparrow$ Stylization & Style
    \end{tabularx}
    \caption{Our method has two mechanisms for controlling the output's level of stylization. In the first row we demonstrate the effect of increasing stylization by ommitting our color post-processing, which increases stylization by allowing greater variation from the hue and chroma of the original content image. In the second row we demonstrate the effect of varying $\alpha$, the parameter controlling the weight of the stylization at the previous scale in the initialization of the next scale. We show results with $\alpha=0.0$ (minimum stylization) and $\alpha=1.0$ (maximum stylization).}
    \label{fig:degree}
\end{figure}

Including or omitting our color processing is an important mechanism for trading off between content preservation and stylization quality (row 1 of Figure \ref{fig:degree}). However, we can also take advantage of our multi-scale procedure to control the stylization level of our final output (row 2 of Figure \ref{fig:degree}).

For both NNST-D and NNST-Opt we produce stylizations at eighth, quarter, half, and full resolution. The upsampled output of the previous scale serves as initialization for the next. We initialize the coarsest scale with a downsampled version of the content image. Let $O_s$ be the output of our algorithm at scale $s$. Let $C_{s+1}, S_{s+1}$ be the content and style images at finer scale $s+1$. Let $O_s^\uparrow$ be $O_s$ upsampled to be the same resolution as $C_{s+1}$. Instead of constructing $T$ by finding matches between $\Phi(C_{s+1})$ and $\Phi(S_{s+1})$, we instead find matches between $\Phi(\alpha O_s^\uparrow + (1-\alpha)C_{s+1})$ and $\Phi(S_{s+1})$. The parameter $\alpha$ controls stylization level, with $\alpha=0$ corresponding to the lowest stylization level, and $\alpha=1$ the highest. By default, we set $\alpha=0.25$, as this generally produces a visually pleasing balance between stylization and content preservation.

%% file: AB3_eval.tex
\definecolor{darkgreen}{rgb}{0.0, 0.5, 0.0}
\definecolor{darkred}{rgb}{0.7, 0.0, 0.0}
\definecolor{darkorange}{rgb}{0.8, 0.6, 0.0}

\begin{table*}[htp]
    \centering
    \begin{tabular}{c|c|c|c|c|}
                  & \multicolumn{4}{c}{Feed-Forward Baselines} \\ \hline
                  & WCT & Avatar & MST & ArtFlow \\ \hline
         NNST-D   & 71\% \small{(1.0e-10)} & 74\% \small{(1.0e-13)} & 70\% \small{(2.5e-10)} & 60\% \small{(1.6e-3)} \\ \hline
         NNST-Opt & 83\% \small{($<$ 1e-15)} & 85\% \small{($<$ 1e-15)} & 72\% \small{(5.9e-12)} & 69\% \small{(1.4e-9)} \\ \hline \hline
        & \multicolumn{4}{c}{Optimization-Based Baselines} \\ \hline
                   &DIA &CNNMRF & Gatys & STROTSS\\ \hline
         NNST-D   & 61\% \small{(4.1e-4)} & 64\% \small{(1.6e-7)} & 60\% \small{(1.0e-3)} & 49\% \small{(0.66)}\\ \hline
         NNST-Opt &  61\% \small{(2.5e-4)} & 82\% \small{($<$1e-15)} & 65\% \small{(1.3e-6)} & 55\% \small{(1.2e-2)}

    \end{tabular}
    \caption{The percentage of votes received by NNST in our forced choice user study when benchmarked against prior work. In parentheses is the p-value of rejecting the null hypothesis that the preference rate for NNST is less than 50\%. Our feed forward variant NNST-D is preferred over all baselines except STROTSS \cite{kolkin2019style}, a much slower optimization based method. Our own optimization based variant, NNST-Opt, is preferred over all baselines}
    \label{tab:study_forced_style}
\end{table*}

\section{Evaluation}
\label{sec:eval}

\subsection{Traditional Media Evaluation Set}
In order to benchmark the performance of our method and prior work we gather a dataset of 30 high-resolution content photographs from Flickr, chosen for their diversity and under the constraint that they be available under a creative commons license allowing modification and redistribution. We followed the same procedure (also using Flickr) to gather ten ink drawings and ten watercolor paintings. From the Riksmuseum's open-source collection we take ten impressionist oil paintings created between 1800-1900. We supplement these with ten pencil drawings taken from the dataset used in Im2Pencil \cite{li2019im2pencil}. In total this gives us 40 high-resolution style images. We use this dataset in the following user study, and will make it available to download.

\subsection{User Study}
In order to assess the stylization quality of NNST-D and NNST-Opt relative prior work we generate stylizations for all pairwise content/style combinations in the traditional media evaluation set described above (A total of 1200 outputs per method). We conduct a user study using Prolific (\url{https://www.prolific.co/}) where users are shown the output of two algorithms (randomly ordered) for the same content/style pair (randomly selected from the 1200 possible combinations), along with the target style, and asked 'Does "Image A" or "Image B" better match the "Target Style"?'. Users are asked to judge 9 such triplets in sequence, among which is mixed one attention verification question (selecting the image that shows a cartoon whale in a randomly ordered triplet). In total we collected 225 votes per method pair, from a total of 400 unique participants. We include 40 comparisons from the user study, along with the content/style inputs, and examples of the study interface in the supplement.

Optimization-based methods are in general considered of higher quality than fast feed-forward ones, therefore each family of techniques are generally compared separately. However, while we group these methods in Table \ref{tab:study_forced_style}, we compare both variants of our method to both families of technique. For optimization-based methods we benchmark against Gatys \cite{gatys2016image}, CNNMRF \cite{li2016combining}, Deep Image Analogies (DIA) \cite{liao2017visual}, and STROTSS~\cite{kolkin2019style}. We were unable to run the official code for CNNMRF and Deep Image Analogies, and re-implemented their methods. For fast feed-forward methods we benchmark against WCT \cite{Li17}, AvatarNet \cite{sheng2018avatar}, MST \cite{zhang2019multimodal}, and ArtFlow \cite{an2021artflow}. 

The results of our study are summarized in Table \ref{tab:study_forced_style}, along with the p-values of rejecting the null hypothesis that the preference rate for NNST-D/Opt is less than 50\%. We calculate these p-values under the assumption that the votes are independent and the sum of votes received by a method is distributed as a binomial. In summary there is a statistically significant preference for our fast variant NNST-D over all benchmarked methods (fast and optimization-based) except STROTSS, a state-of-the-art optimization-based method, for which NNST-D is on par with (STROTSS is preferred but not by a statistically significant margin). There is a statistically significant preference for NNST-Opt over {\em all} benchmarked methods.

\subsection{Discussion} 

Our approach performs well in general but there is nonetheless areas where there remains room for improvement. For instance, physical phenomena like the drips of paint on the bear in Figure~\ref{fig:face} are not reproduced. Also lines and hatching patterns like in Figure~\ref{fig:qual_fast} (second and last rows) can be distorted. These cases are challenging for all methods, and while our approach often better reproduce these phenomena and patterns than previous work, we believe that further exploring this direction would be a worthwhile effort in the future.

We also observe that even though the decoder variant produces good results, they are not as good as the optimization-based reconstructions. Reducing that gap will be key to creating a high-quality practical stylization algorithm.

\begin{figure}
\includegraphics[width=\linewidth]{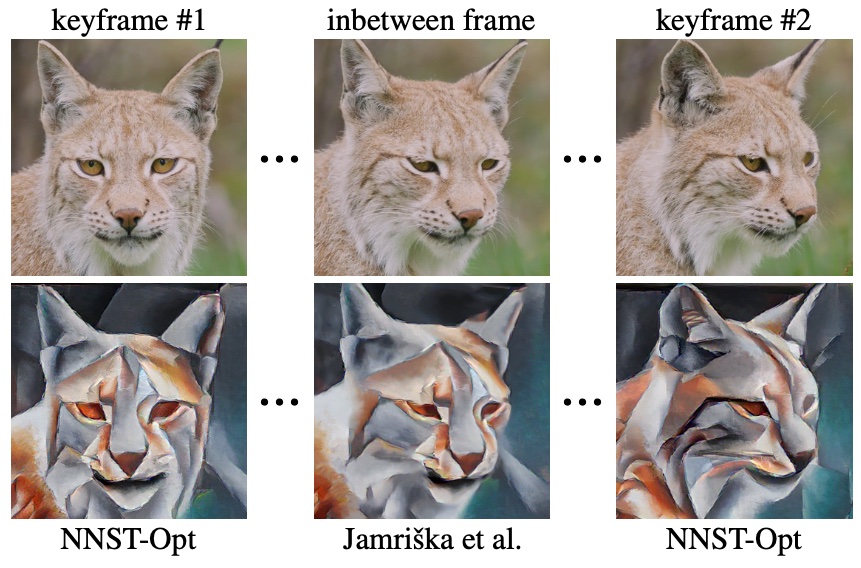}
\caption{Our approach used as a generator of stylized frames
for example-based video stylization---a few selected keyframes are stylized using NNST-Opt and the rest of the sequence is stylized using the method of Jamri\v{s}ka et al.~\cite{Jamriska19}. See video \href{https://home.ttic.edu/~nickkolkin/nnst_video_supp.mp4}{here}.}
\label{fig:lynx}
\end{figure}

\begin{figure} 
\includegraphics[width=\linewidth]{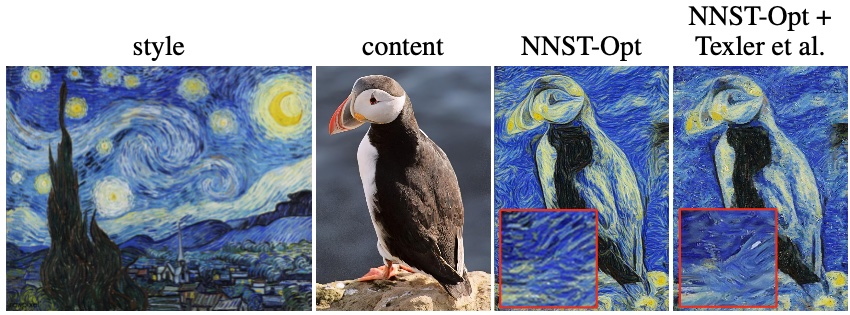}
\caption{Our approach combined with the method of Texler et al.~\cite{Texler20-CAG}---the output of NNST-Opt is used as a
guide for patch-based synthesis algorithm that can operate at notably higher
resolution, in this case a 4K resolution output, and thus faithfully preserve high-frequency details (see the zoom-in insets in red squares). At the mid-scale level, however, our technique still
outperforms patch-based synthesis as it can convey style features better to delineate salient structures in the content image.} \label{fig:puffin}
\end{figure}

\section{Extensions and Applications}

\begin{figure}
\includegraphics[width=\linewidth]{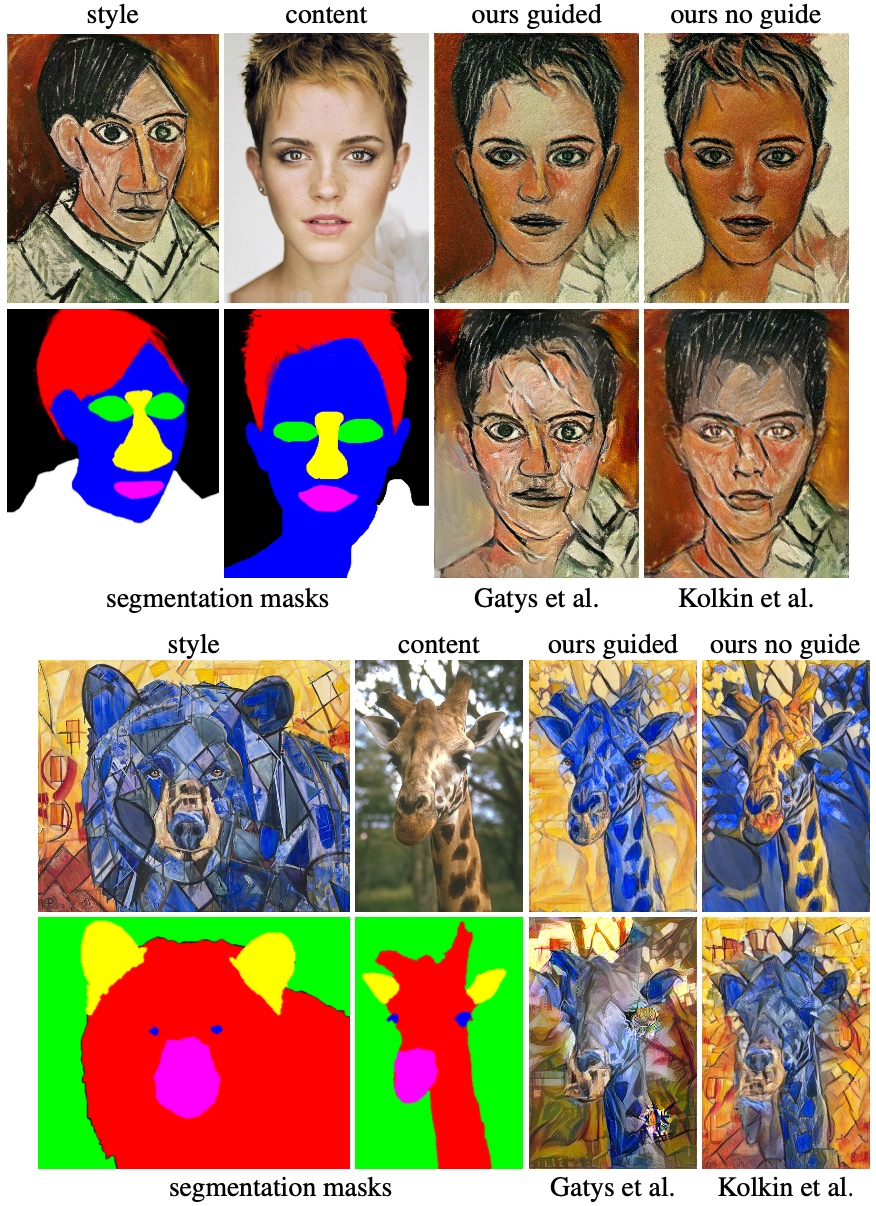}
\caption{Incorporating additional guidance (segmentation masks) into our technique---in contrast to unguided version the output
is semantically meaningful, i.e., background texture in the stylized image corresponds to the background in the style exemplar, etc. When compared to the current state-of-the-art in neural style transfer that also support guidance~\cite{gatys2017controlling, kolkin2019style} our approach better preserves style details.}\label{fig:face}
\end{figure}

Although the primary goal of our approach is to perform a generic style transfer
without requiring additional knowledge about the style and content image, in
the case when such information is available our technique can be easily extended to incorporate it. In this scenario we follow the concept of Image Analogies~\cite{hertzmann2001image} and extend our objective~(\ref{eq:match}) by
adding a term that incorporates further guidance on top of the cosine distance:
\begin{equation*}\label{eq:guided}
\begin{split}
    T_i = \arg\min_{\Phi'(S)_j} w_{\cos}D\Big(\Phi(C)_i - \mu_C, \hspace{0.2cm}\Phi'(S)_j - \mu'_S \Big) \\ + w_{guide}D^g(C^g_i, S^g_j)
\end{split}
\end{equation*}
Here~$S^g_i$ and~$C^g_i$ are downsampled versions of style and content guiding
channels (e.g., segmentation masks, see Figure~\ref{fig:face}), $D^g$~is a
metric which evaluates guide similarity at pixels~$i$ and~$j$ (in our
experiments we use sum of squared differences), and~$w_{\cos}$ and~$w_{guide}$
are weights that balance the influence of the cosine and guiding term (in our experiments we set~$w_{\cos}=0.5$ and~$w_{guide}=0.5$). In
Figure~\ref{fig:face} we demonstrate the effect of incorporating additional
segmentation masks as a guiding channels. It is visible that with guidance the
content from the style is transferred in a more semantically meaningful way. In
contrast to previous neural approaches that also support guidance~\cite{gatys2017controlling, kolkin2019style} our technique better preserves visual aspects of the original
style exemplar.

Besides single image style transfer our approach is practical also in the context of example-based video stylization~\cite{Jamriska19,Texler20-SIG} where the aim is to propagate the
style from a sparse set of stylized keyframes to the rest of the video sequence. In
the original setting, the stylization of keyframes is tedious as those need to
be created by hand to stay perfectly aligned with the content in the
video. Using our approach, however, one can stylize the entire sequence
fully automatically without the need to preserve alignment. By transferring
the style from an arbitrary exemplar image one can stylize a subset of frames
and then run an existing keyframe-based video stylization technique of
Jamri\v{s}ka et al.~\shortcite{Jamriska19} or Texler et al.~\shortcite{Texler20-SIG} to propagate the style to the rest of the sequence while maintaining temporal coherence (see~Figure~\ref{fig:lynx} and
our supplementary video).

One of the limiting factors of our technique is that using currently available
GPUs it can deliver only outputs in moderate resolutions such as 1k. To
obtain higher resolution images our technique can be plugged into the method of
Texler et al.~\shortcite{Texler20-CAG}. In this approach the result of neural style
transfer is used as a guide to drive patch-based synthesis
algorithm of Fi\v{s}er et al.~\shortcite{Fiser16} that can produce a high-resolution counterpart of the stylized image generated by the neural method (in our case the generated nearest neighbor field is upsampled to obtain a 4K output). Such a combination can help to ensure a more faithful
style transfer thanks to the ability to reproduce high-frequency details of the
original style exemplar. However, a compromise here is that when comparing
middle scale features our technique can perform better than patch-based
synthesis since it can adopt the style features to follow salient structures visible in the content image (c.f.~Figure~\ref{fig:puffin}).

%% file: AB4_conclusion.tex
\section{Conclusion}

We have demonstrated a conceptually simple approach to artistic stylization of images. We explored several key design choices to motivate our algorithm. We showed qualitatively and quantitatively that our approach is flexible enough to support various scenarios and produce high-quality results in all these cases. Put together, we believe that these characteristics make our approach suitable for practical applications and a solid basis for future work. 




%% file: AB5_Design.tex
\section{Supplement - Design Decisions}

A pithy description of NNST and the non-parametric neural style transfer algorithm proposed by Chen and Schmidt\cite{chen2016fast} much earlier in 2016 would reveal little difference between the two. Both methods explicitly construct a tensor of `target features' by replacing vectors of VGG-derived content features with vectors of VGG-derived style features, then optimize the pixels of the output image to produce the 'target features' (or use a learned decoder). Yet, there is a dramatic difference between the visual quality of the algorithms' outputs. As is often the case, the devil is in the details, and this section explores the important design decisions that can boost a style transfer algorithm's visual quality.

\begin{figure*}[htp]
    \centering
    \includegraphics[width=\linewidth]{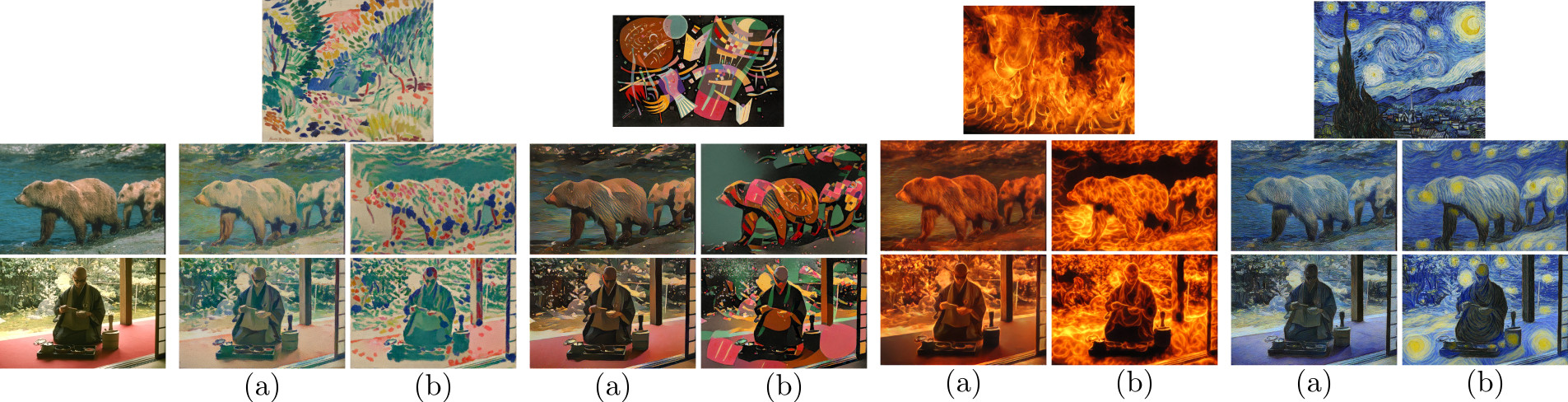}
    \caption{Visual comparison between (a.) the outputs of Chen and Schmidt \cite{chen2016fast} and (b.) a simplified variant of NNST which uses the feature splitting regime across all scales and does not employ color correction. While both algorithms share a similar high level framework, they differ in many details, resulting in NNST much better recreating distinctive visual features of the style image.}
    \label{fig:chen_nnst_comp}
\end{figure*}

\begin{figure*}[htp]
    \centering
    \includegraphics[width=\linewidth]{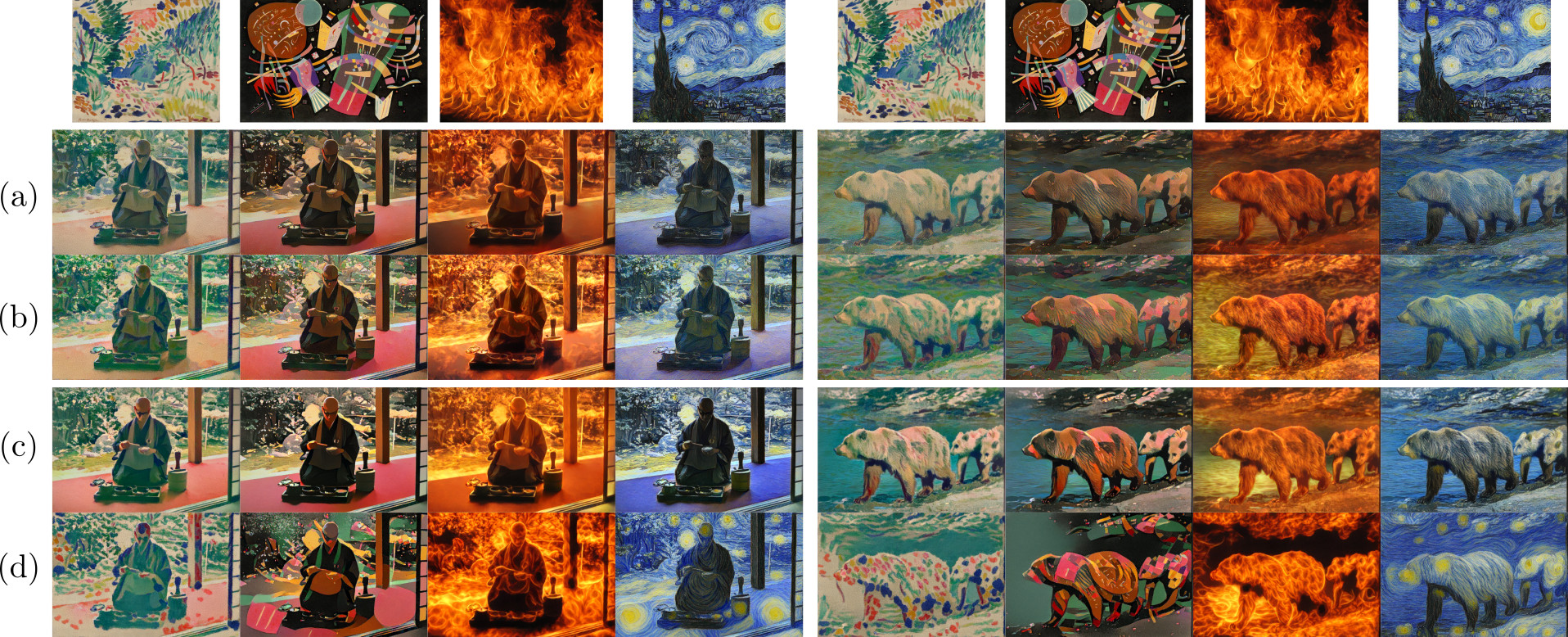} 
    \caption{Visual comparison between matching features separately for each location ($1\times1$ patches) and matching them as $3\times3$ patches: (a.) the outputs of Chen and Schmidt \cite{chen2016fast} (3x3 feature patches matched and overlaps averaged), (b.) a variant of \cite{chen2016fast} where feature patches are matched independently for each spatial location (1x1 patches, no averaging), (c.) a variant of NNST where 3x3 patches are matched and averaged, and (d.) the simplified NNST variant from Figure \ref{fig:chen_nnst_comp}. Note that matching 1x1 rather than 3x3 patches (b. and d. relative to a. and c.) allows more high frequency details of the style to appear in the output.}
    \label{fig:chen_patch}
\end{figure*}

\begin{figure*}[htp]
    \centering
    \includegraphics[width=\linewidth]{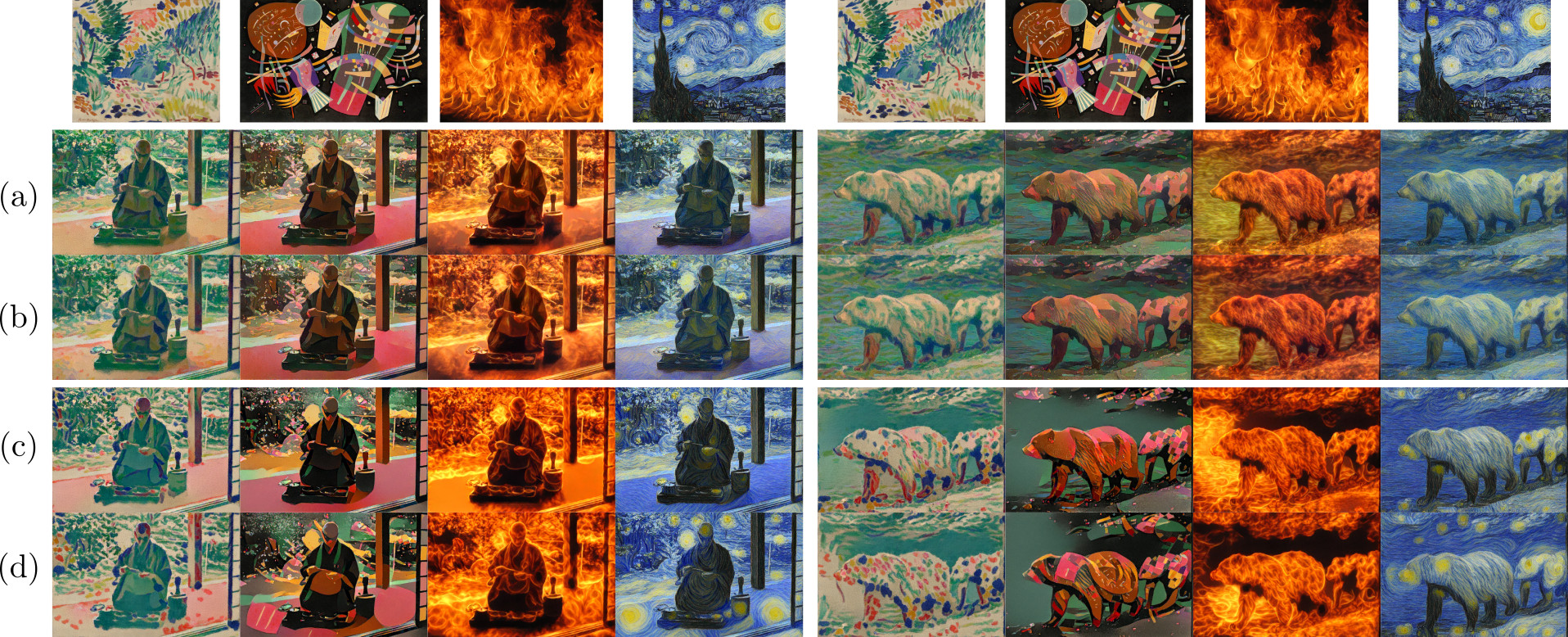} 
    \caption{Visual comparison between single-scale and multi-scale stylization: (a.) \cite{chen2016fast} w/ 1x1 patches (row b of Figure \ref{fig:chen_patch}), (b.) a. applied coarse-to-fine using the same mechanisms as NNST ($\alpha=0.25$), (c) simplified NNST at only the finest scale ($\alpha=1.0$), and (d.) the simplified NNST variant from Figure \ref{fig:chen_nnst_comp}. Stylizing coarse-to-fine increases stylization level and results in visual features of the style with larger spatial extent appearing in the output (and this effect increases with lower $\alpha$, see Figure \ref{fig:degree}). In addition, stylizing coarse-to-fine allows stylistic details to be hallucinated in large flat regions of the content image (compare the floor beneath the monk in c. and d.).}
    \label{fig:chen_scale}
\end{figure*}

\begin{figure*}[htp]
    \centering
    \includegraphics[width=\linewidth]{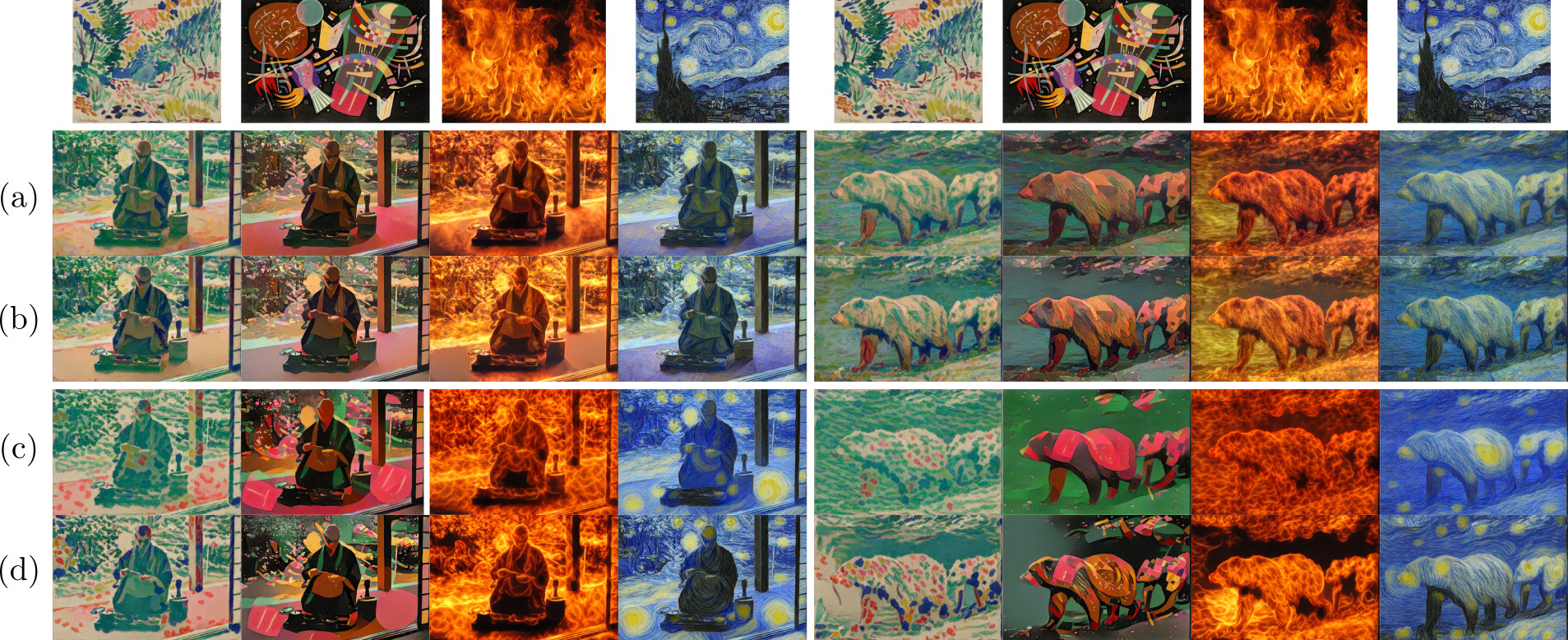} 
    \caption{Visual comparison between using zero-centering or not before matching features using the cosine distance: (a.) multi-scale \cite{chen2016fast} w/ 1x1 patches (row b of Figure \ref{fig:chen_scale}), (b.) a. using the centered cosine distance for feature matching (instead of the standard cosine distance), (c) d. using the standard cosine distance (instead of the centered cosine distance), and (d.) the simplified NNST variant from Figure \ref{fig:chen_nnst_comp}. Using the centered cosine distance not only results in a more diverse set of style features appearing in the output, the contrast between these features helps preserve the perceived contents of the original input.}
    \label{fig:chen_metric}
\end{figure*}

\begin{figure*}[htp]
    \centering
    \includegraphics[width=\linewidth]{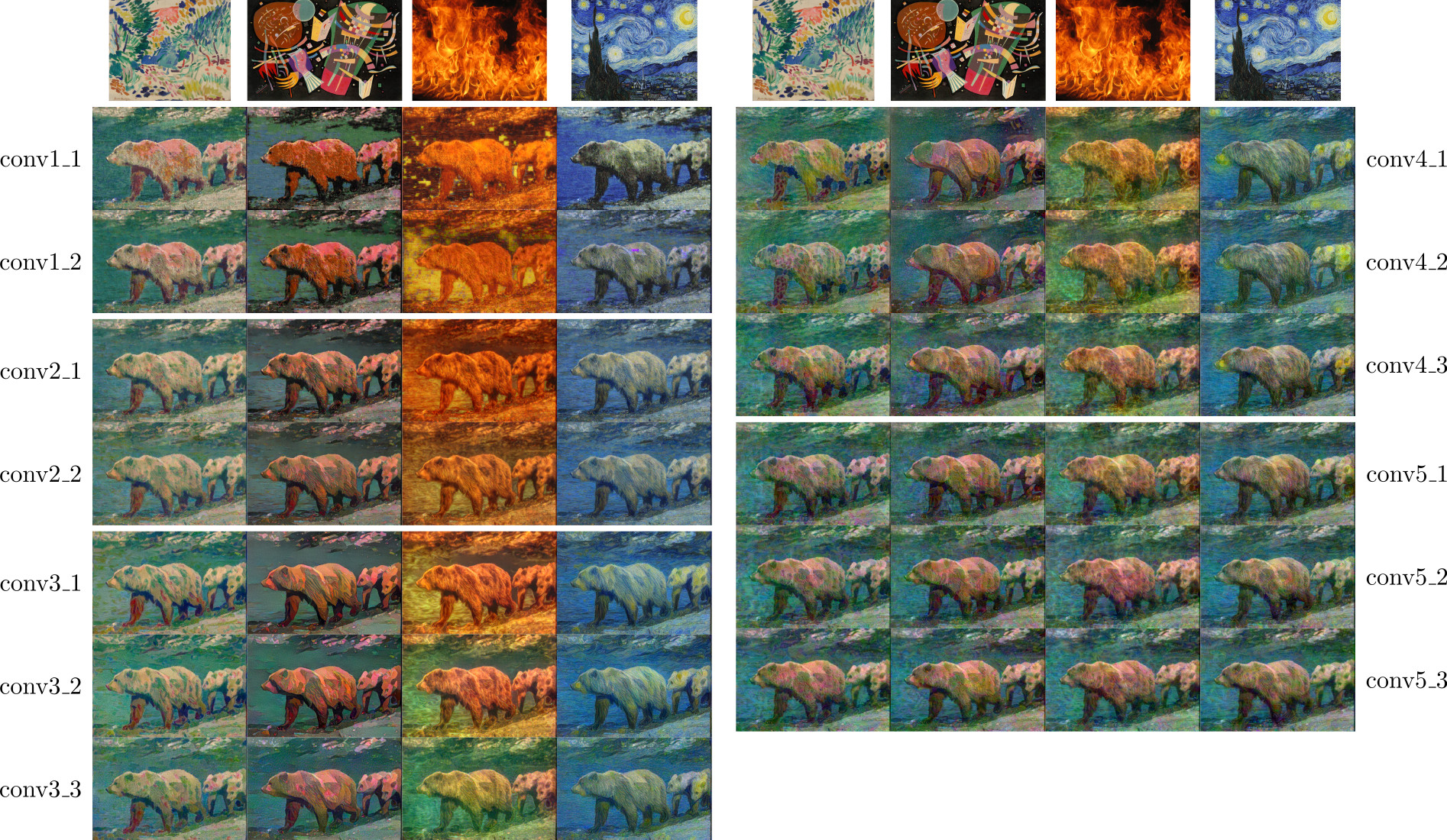} 
    \vspace{0.2cm}
    \includegraphics[width=\linewidth]{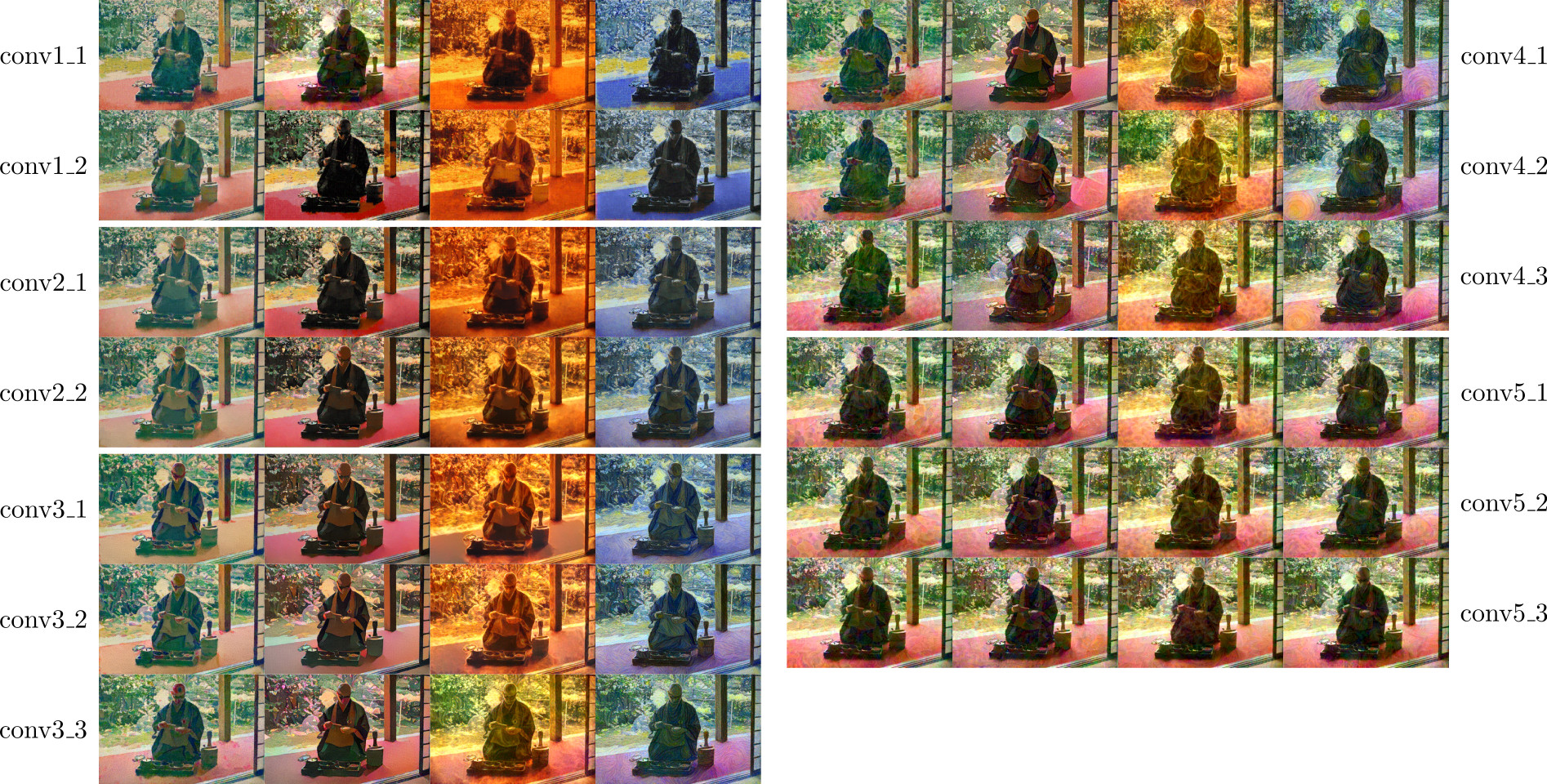}
    \caption{Visual comparison between using different individual convolutional layers of pretrained VGG-16 as a source of features. All images are produced using multi-scale \cite{chen2016fast} w/ 1x1 patches matched with the centered cosine distance (row b. of Figure \ref{fig:chen_metric} corresponds to row conv3\_1 of this figure, the default style features used by \cite{chen2016fast}). Layers in the first two conv. blocks capture color well, but not more complex visual elements. Layers in the third and fourth conv block capture complex visual elements, but not color. Layers in the fifth block do not seem closely tied to stylistic features. No layer alone is sufficient to capture all desired stylistic features.}
    \label{fig:chen_fsingle}
\end{figure*}

\begin{figure*}[htp]
    \centering
    \includegraphics[width=\linewidth]{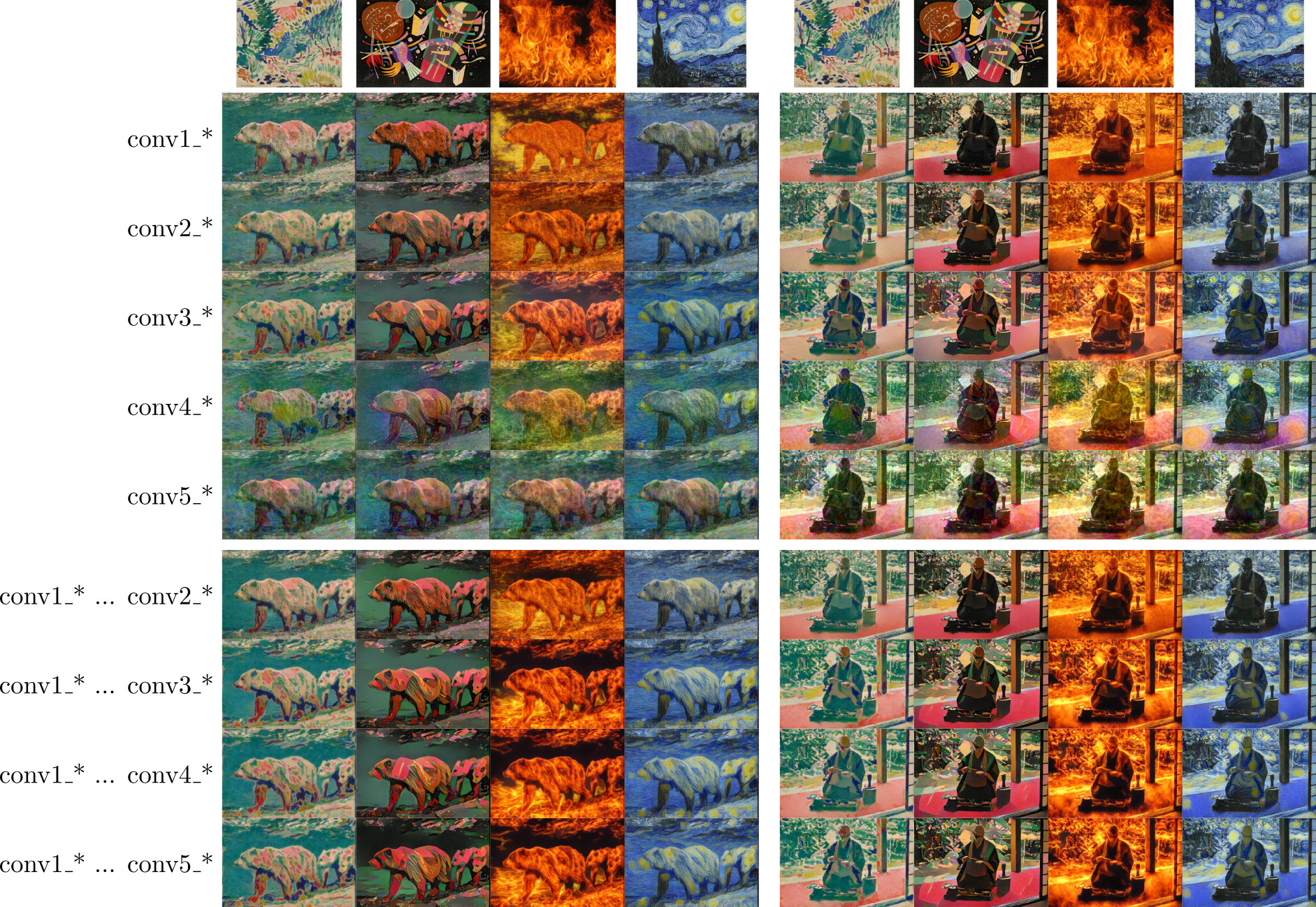} 
    \caption{Visual comparison between using features from multiple layers of pretrained VGG-16. The first 5 rows demonstrate the effect of using all the layers from a given conv. block. Just as no single layer is sufficient, no single conv. block contains a rich enough representation of style to produce satisfactory outputs. The 5th-9th rows demonstrate the effect of using all of the features up to a certain depth in the network. Most important stylistic details can be captured using the first three conv. blocks. Small improvements can be made using the 4th and 5th conv. blocks as well, but it is probably not worth the computational cost (the 4th and 5th blocks each contain 36\% of the total feature channels). While NNST uses all feature through conv. block 4, only using features through block 3 would be an obvious means to increase efficiency.}
    \label{fig:chen_fblock}
\end{figure*}

\begin{figure*}[htp]
    \centering
    \includegraphics[width=\linewidth]{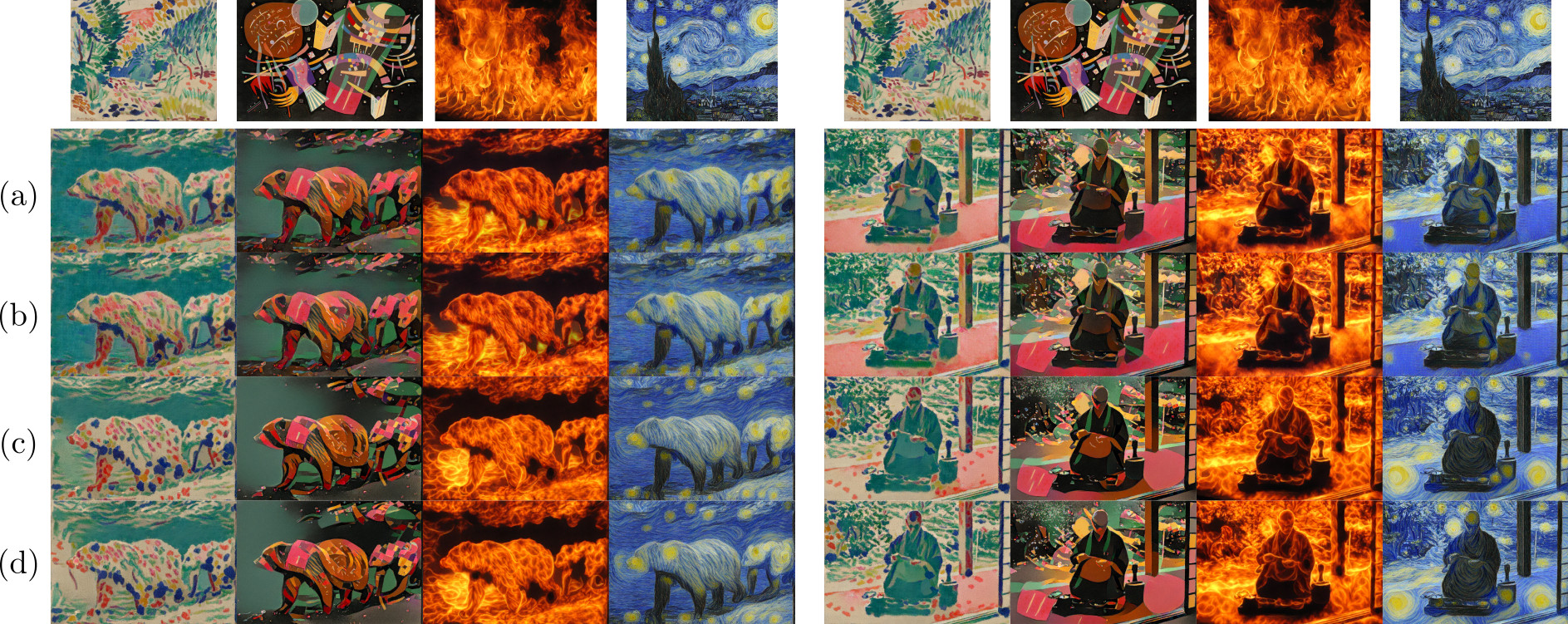} 
    \caption{The design decisions so far lead to an algorithm close to NNST, the only difference that remain to be evaluated are the frequency of computing matches and whether or not to compute matches separately for each layer: (a.) multi-scale \cite{chen2016fast} w/ 1x1 hypercolumns using layers conv1\_1-conv4\_3, matched with the centered cosine distance (row 8 of Figure \ref{fig:chen_fblock}), (b.) a. with nearest neighbors recomputed after each update of the output image, (c.) a. with nearest neighbors computed seperately for each layer, and (d.) the simplified NNST variant from Figure \ref{fig:chen_nnst_comp} (i.e. a. w/ features matched separately for each layer and recomputing matches each update). Nice results can already be achieved without feature splitting or recomputing matches each iteration (a.), however slightly sharper high frequencies can be achieved by recomputing matches (b.), and more diverse stylistic features from each layer are matched seperately (c.). When both of these modifications are applied, we essentially arrive at NNST (d.).}
    \label{fig:chen_freq_split}
\end{figure*}

In Figures \ref{fig:chen_nnst_comp}, \ref{fig:chen_patch}, \ref{fig:chen_scale}, \ref{fig:chen_metric}, \ref{fig:chen_fsingle}, \ref{fig:chen_fblock}, \ref{fig:chen_freq_split}, we visually explore the effects of the our design decisions relative to \cite{chen2016fast}, and iteratively modify their method until arriving at NNST. Where appropriate these figures also demonstrate the effect of modifying individual design elements of NNST to match \cite{chen2016fast}.

\newpage